\documentclass[10pt,journal,compsoc]{IEEEtran}
\usepackage{amsmath,amssymb} 
\usepackage{mathrsfs}
\usepackage{multirow}
\usepackage{multicol}
\usepackage{hyperref}
\usepackage{url}
\usepackage{amsthm}
\usepackage{graphicx}
\usepackage{booktabs}
\usepackage{multirow}
\usepackage{amsfonts}
\usepackage{bbm}
\usepackage[normalem]{ulem}
\usepackage{bm}
\usepackage{subfigure}
\usepackage{siunitx}
\usepackage{balance}  
\useunder{\uline}{\ul}{}

\def\eg{e.g.}
\def\ie{i.e.}

\newcommand{\secref}[1]{section~\ref{sec:#1}}

\newcommand{\eqnref}[1]{Eqn.~\ref{#1}}
\newcommand{\firef}[1]{Fig.~\ref{#1}}

\newcommand{\ParagraphB}[1]{\noindent\textbf{#1}}
\renewcommand{\raggedright}{\leftskip=0pt \rightskip=0pt plus 0cm}

\usepackage{tikz}

\usepackage{caption}
\usepackage{xcolor}
\usepackage[disable]{todonotes}
\usepackage{pifont}
\tikzstyle{edge}=[-latex',draw=black!90,shorten <=1pt,shorten >=1pt]
\tikzstyle{redge}=[latex'-,draw=black!90,shorten <=1pt,shorten >=1pt]
\tikzstyle{dedge}=[latex'-latex',draw=black!90,shorten <=1pt,shorten >=1pt]

\tikzstyle{block}=[draw, text width=5em,align=center,shape=rectangle, rounded corners, , align=center]
\tikzstyle{nobox}=[align=center]
\definecolor{emb}{RGB}{209,228,252}
\definecolor{hidden-blue}{RGB}{194,232,247}
\usetikzlibrary{fadings}
\usetikzlibrary{mindmap,trees}
\usetikzlibrary{arrows,automata,shapes,positioning,shadows,trees}
\usetikzlibrary{shapes,snakes,shadows}
\usetikzlibrary{shapes.arrows}
\usetikzlibrary{calc,shapes, positioning}
\usetikzlibrary{decorations.text}
\usetikzlibrary{matrix,chains,positioning,decorations.pathreplacing,arrows}
\usetikzlibrary{bayesnet}
\usepackage[edges]{forest}
\usetikzlibrary{shadows.blur}
\usetikzlibrary{shapes.geometric}
\definecolor{myred}{HTML}{F8F6F4} 
\definecolor{mypurple}{HTML}{FFDFDF}
\definecolor{myyellow}{HTML}{FFF6F6}
\definecolor{mygreen}{HTML}{D2E0FB} 

\usetikzlibrary{arrows,shapes,positioning,shadows,trees}

\tikzset{
	basic/.style  = {draw, text width=2cm, drop shadow, font=\sffamily, rectangle},
	root/.style   = {basic, rounded corners=2pt, thin, align=center,
		fill=green!30},
	level 2/.style = {basic, rounded corners=6pt, thin,align=center, fill=green!60,
		text width=8em},
	level 3/.style = {basic, thin, align=left, fill=pink!60, text width=6.5em}
}

\tikzstyle{leaf1}=[draw=black, 
rounded corners,minimum height=1.2em,
edge=black!10, 
text opacity=1, align=center,
fill opacity=.3,  text=black,font=\scriptsize,
inner xsep=2pt, inner ysep=3.6pt,
]
\tikzstyle{leaf2}=[draw=black, 
rounded corners,minimum height=1.2em,
edge=black!10, 
text opacity=1, align=center,
fill opacity=.5,  text=black,font=\scriptsize,
inner xsep=2pt, inner ysep=3.6pt,
]
\tikzstyle{leaf3}=[draw=black, 
rounded corners,minimum height=1.2em,
edge=black!10, 
text opacity=1, align=center,
fill opacity=.8,  text=black,font=\scriptsize,
inner xsep=2pt, inner ysep=3.8pt,
]
\tikzstyle{leaf4}=[draw=black, 
rounded corners,minimum height=1.2em,
text width=4.4em, 
edge=black!10, 
text opacity=1, align=center,
fill opacity=1,  text=black,font=\scriptsize,
inner xsep=2pt, inner ysep=3.8pt,
]
\AtBeginDocument{%
	\providecommand\BibTeX{{%
			\normalfont B\kern-0.5em{\scshape i\kern-0.25em b}\kern-0.8em\TeX}}}

\ifCLASSOPTIONcompsoc
\usepackage[nocompress]{cite}
\else
\usepackage{cite}
\fi
\ifCLASSINFOpdf
\else
\fi

\begin{document}

\title{A Comprehensive Survey on Evidential Deep Learning and Its Applications}

\author{Junyu~Gao,
	Mengyuan Chen,
	Liangyu Xiang,
	and~Changsheng~Xu,~\IEEEmembership{Fellow,~IEEE}
	\IEEEcompsocitemizethanks{\IEEEcompsocthanksitem Junyu Gao, Mengyuan Chen,
		Liangyu Xiang, and Changsheng Xu are with State Key Laboratory of Multimodal Artificial Intelligence Systems, Institute of Automation, Chinese Academy of Sciences, Beijing 100190, P. R. China, and with School of Artificial Intelligence, University of Chinese Academy of Sciences, Beijing, China. Changsheng Xu is also with Peng Cheng Laboratory, ShenZhen 518055, China. (e-mail: \{junyu.gao, csxu\}@nlpr.ia.ac.cn, yaoxuan2022@ia.ac.cn).}
	\thanks{Manuscript received April 19, 2005; revised August 26, 2015.}}

\markboth{Journal of \LaTeX\ Class Files,~Vol.~14, No.~8, August~2015}%
{Shell \MakeLowercase{\textit{et al.}}: Bare Demo of IEEEtran.cls for Computer Society Journals}

\IEEEtitleabstractindextext{%
\begin{abstract}
\raggedright{
Reliable uncertainty estimation has become a crucial requirement for the industrial deployment of deep learning algorithms, particularly in high-risk applications such as autonomous driving and medical diagnosis.
However, mainstream uncertainty estimation methods, based on deep ensembling or {Bayesian neural networks}, generally impose substantial computational overhead.
To address this challenge, a novel paradigm called Evidential Deep Learning (EDL) has emerged, providing reliable uncertainty estimation with minimal additional computation in a single forward pass.
This survey provides a comprehensive overview of the current research on EDL, designed to offer readers a broad introduction to the field without assuming prior knowledge.
Specifically, we first delve into the theoretical foundation of EDL, the subjective logic theory, and discuss its distinctions from other uncertainty estimation frameworks.
We further present existing theoretical advancements in EDL from four perspectives: reformulating the evidence collection process, improving uncertainty estimation via OOD samples, delving into various training strategies, and evidential regression networks.
Thereafter, we elaborate on its extensive applications across various machine learning paradigms and downstream tasks.
In the end, an outlook on future directions for better performances and broader adoption of EDL is provided, highlighting potential research avenues.}
\end{abstract}

%

}


\maketitle

\IEEEdisplaynontitleabstractindextext

%
\IEEEpeerreviewmaketitle

\IEEEraisesectionheading{\section{Introduction}}

\IEEEPARstart{O}{ver} the past decade, deep learning has brought revolutionary changes to the field of artificial intelligence~\cite{ghahramani2015probabilistic,pereira2021interactomes}.
Thanks to effective training techniques such as Dropout~\cite{srivastava2014dropout} and residual connections~\cite{he2016deep}, along with outstanding network architectures like Transformer~\cite{vaswani2017attention}, modern neural networks have achieved unprecedented success across almost all applications of machine learning~\cite{zou2023object,li2014secrets}.
However, the expanding range of real-world applications, particularly in high-risk areas such as autonomous driving~\cite{choi2019gaussian}, medical diagnosis~\cite{razzak2018deep}, and military applications~\cite{svenmarck2018possibilities}, has raised increasing demands for the safety and interpretability of neural networks.
Consequently, reliable uncertainty estimation has become a crucial and hotly debated topic in deep learning~\cite{ghahramani2015probabilistic,gal2016dropout,ritter2018scalable,rahaman2021uncertainty,sensoy2018evidential,guo2022survey}.

\begin{figure}[t!]
	\centering
	\includegraphics[width=1\linewidth]{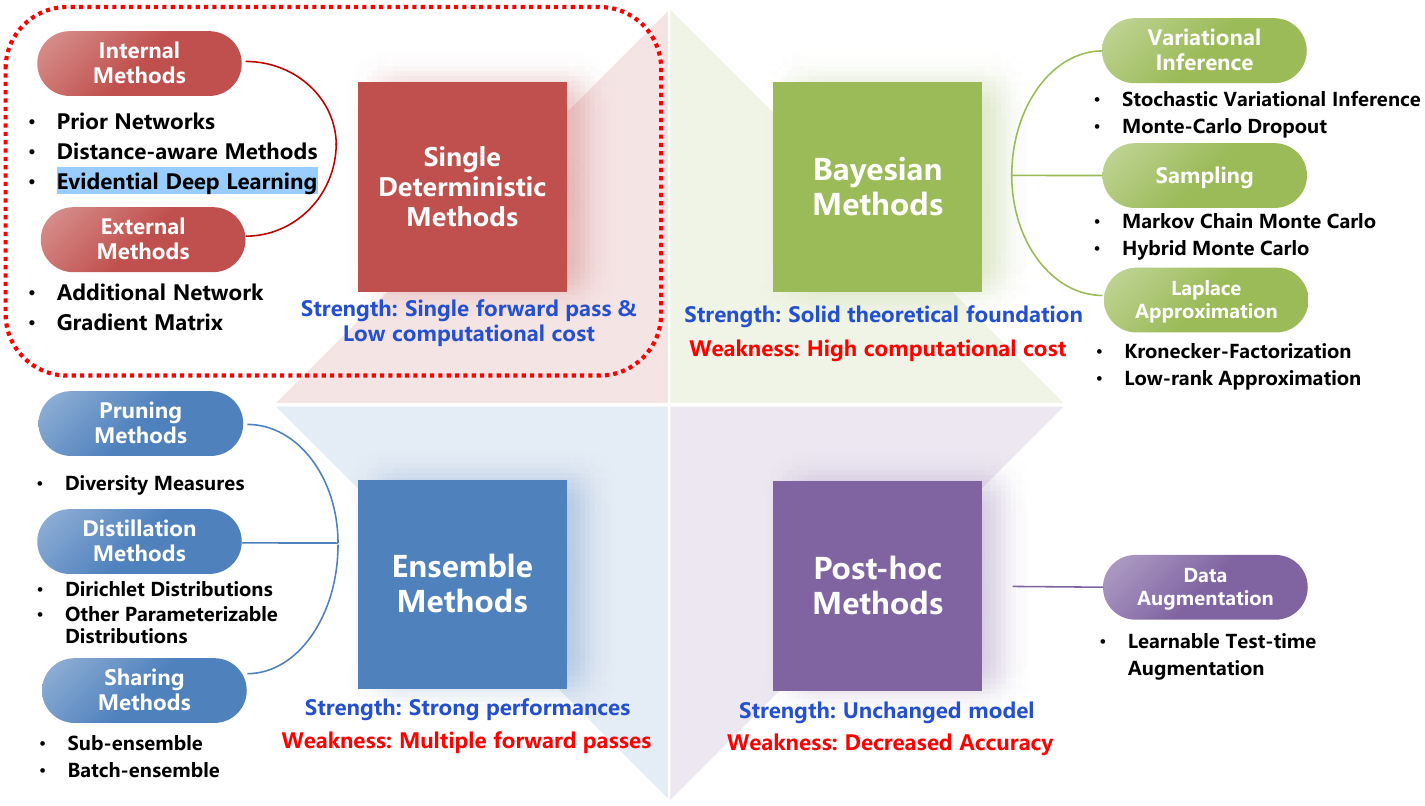}
	\caption{Main categories of existing uncertainty quantification methods, including: single deterministic methods, Bayesian methods, ensemble methods, and post-hoc methods.}
	\label{related}
	\vspace{-4mm}
\end{figure}

As shown in \firef{related}, the mainstream uncertainty quantification methods, such as deep ensembling~\cite{lakshminarayanan2017simple,wen2020batchensemble} and {Bayesian neural networks}~\cite{neal2012bayesian,gal2016dropout,izmailov2021bayesian}, generally involve multiple forward passes or additional parameters, imposing substantial computational burdens that impede their widespread industrial adoption.
To side-step this conundrum, a newly arising single-forward-pass uncertainty estimation paradigm which obtains reliable uncertainty with minimal additional computation, namely Evidential Deep Learning (EDL)~\cite{sensoy2018evidential}, has been extensively developed.
EDL explores the subjective logic theory~\cite{josang2001logic,josang2016subjective}, an advanced variant of the well-known Dempster-Shafer Evidence Theory~\cite{shafer1992dempster,dempster2008upper}, in the domain of deep neural networks. Unlike traditional probabilistic logic, subjective logic assigns belief masses to possible categories to represent the truth of propositions and explicitly includes uncertainty mass to convey the meaning of "I don't know" or "I'm indifferent."
To model the posterior predictive distribution, a Dirichlet distribution is constructed, which is bijective to a subjective opinion comprising the aforementioned belief masses and uncertainty mass. As the uncertainty mass increases, the Dirichlet distribution gradually reverts to a preset prior distribution.
Utilizing these properties of subjective logic, EDL employs a deep neural network as an evidence collector to generate appropriate belief masses and uncertainty mass. The model optimization is then performed by minimizing traditional loss functions integrated over the corresponding Dirichlet distribution~\cite{sensoy2018evidential,chen2023r,amini2020deep}.

This survey delivers an in-depth exploration of the latest developments in Evidential Deep Learning, aiming to introduce readers to the field comprehensively without assuming prior familiarity. We begin by introducing the theory of EDL, focusing on the principles of subjective logic theory and contrasting it with other frameworks for uncertainty estimation. This is followed by a detailed discussion of recent theoretical advancements in EDL, organized into four key areas: reformulating the evidence collection process, improving uncertainty estimation via OOD samples, delving into various training strategies, and evidential regression networks. Additionally, we illustrate the broad applicability of EDL across different machine learning paradigms and various downstream tasks. The survey concludes with a forward-looking perspective 
aimed at enhancing EDL's capabilities and facilitating its wider applications, pinpointing promising areas for further investigation.

Note that two surveys~\cite{ulmer2021prior,cerutti2022evidential} related to evidential learning have been published in the current field, compared to them, the primary distinctions of our work are as follows:
(1)~\cite{ulmer2021prior} provides a broad overview of a series of methods for parameterizing Dirichlet priors or posterior distributions and~\cite{cerutti2022evidential} focuses on
evidential reasoning and learning from the process
of Bayesian update of given hypotheses based on
additional evidence. Different from the two surveys, which primarily focus on formulations also widely adopted by other machine learning paradigms (\eg,~priors or posterior distributions, Bayesian learning), we specifically focus on the EDL method itself and its recent theoretical advancements;
(2)~we delve into several key concepts from the intrinsically theoretical foundation of EDL, the subjective logic theory, and additionally illustrate the connections and differences between subjective logic and other uncertainty reasoning frameworks.
However,~\cite{ulmer2021prior,cerutti2022evidential} lacks attention to the construction process of evidence learning guided by subjective logic theory;
(3)~compared with the two surveys, we additionally and comprehensively conclude the extensive applications of EDL across various machine learning paradigms and downstream tasks.


The following content is organized as follows: we begin by providing an overview of existing uncertainty quantification techniques in deep learning, including single deterministic methods, Bayesian methods, ensemble methods, {and post-hoc methods} (\secref{related}).
Next, we delve into the theoretical basis of EDL, specifically the subjective logic theory (\secref{subjective}), and compare it with other uncertain reasoning frameworks (\secref{comparison}).
We then focus on the mechanism of the standard EDL method (\secref{evidential}) and explore several theoretical advancements around: reformulating the evidence collection process (\secref{reformulating}), improving uncertainty estimation with out-of-distribution (OOD) samples (\secref{improving}), investigating various training strategies (\secref{delving}), and deep evidential regression (\secref{regression}).
Additionally, we introduce the EDL enhanced machine learning (\secref{machine}) and EDL in downstream applications (\secref{downstream}).
Finally, we present our conclusions and discuss potential future directions in \secref{conclusion}. A comprehensive list of
EDL methods and its applications can be found at \url{https://github.com/MengyuanChen21/Awesome-Evidential-Deep-Learning}.

\vspace{-1mm}
\section{Related Works}
\label{sec:related}

In this section, we begin by explaining some common terms of uncertainty categories (\secref{term}), and then elaborate on four main categories of existing uncertainty estimation methods (\secref{uqm}), including single deterministic methods, Bayesian methods, ensemble methods, and post-hoc methods.
We believe that the elucidation of these terms and methods is crucial for establishing a clear understanding of the landscape of uncertainty quantification in deep learning.

\vspace{-1mm}
\subsection{{Common Terms of Uncertainty Categories}}
\label{sec:term}

\textbf{Epistemic and Aleatoric Uncertainties.} A prevalent classification of uncertainty sources divides them into epistemic and aleatoric categories~\cite{der2009aleatory,guo2022survey}.
\textbf{Epistemic uncertainty}, also known as \textbf{model uncertainty}, stems from insufficient knowledge or limited data.
High epistemic uncertainty suggests that the model does not possess enough information to make a reliable prediction for a given sample.
This often occurs when the sample is significantly different from the training data, indicating the presence of an out-of-distribution (OOD) sample.
In contrast, \textbf{aleatoric uncertainty}, or \textbf{data uncertainty}, is inherent to the nature of the training data itself.
This type of uncertainty can arise from various factors, such as the intrinsic similarity between certain classes (e.g., digits 0 and 6), noise introduced during data collection, and inaccurate or erroneous annotations. 
Aleatoric uncertainty persists regardless of the amount of additional data or improvements made to the model, representing the irreducible variability in the data that limits the precision of predictions.
Generally speaking, epistemic uncertainty can be reduced by gathering more data and enhancing the model, while aleatoric uncertainty requires careful consideration of the data's inherent properties and may necessitate alternative approaches to improve prediction reliability in the presence of such variability.
Besides, current discussions~\cite{zhao2019quantifying,chen2023uncertainty} predominantly suggest that the uncertainty estimated by the vanilla EDL method pertains to epistemic uncertainty.

\noindent\textbf{Vacuity and Dissonance.} In the field of uncertainty estimation, the concepts of \textbf{vacuity} and \textbf{dissonance} are also common terms adopted to describe uncertainties from different sources.
\textbf{Vacuity} refers to the uncertainty arising from a lack of information or knowledge, often synonymous with epistemic uncertainty.
In contrast, \textbf{dissonance} describes uncertainty arising from conflicting sources of information.
This typically occurs when a model gives inconsistent predictions based on different parts or features of a particular sample.
Even though the model might have sufficient information for individual features, the overall uncertainty increases due to the conflict between these pieces of information. 
Dissonance does not directly correspond to aleatoric or epistemic uncertainty, but it can be seen as a complex form of uncertainty containing elements of both.
\cite{zhao2019quantifying} provides a formulation of dissonance in the framework of EDL, whose details are introduced in \secref{improving}.

\vspace{-1mm}
\subsection{Existing Uncertainty Quantification Methods}
\label{sec:uqm}
\textbf{Single deterministic methods.} A deterministic neural network refers to a network where all parameters are deterministic, meaning that repeating a forward pass multiple times will yield same results. 
In the field of uncertainty quantification, single deterministic methods generally refer to a series of algorithms that can estimate uncertainty with a single forward pass on a deterministic network, whose advantage is that they are computationally much more efficient than most Bayesian methods and ensemble methods.
Single deterministic methods can be further categorized into internal and external methods based on whether they use additional components to quantify uncertainty, as shown in \firef{related}.
(1) \textit{Internal methods} generally train a single forward network to generate the parameters of a predictive distribution, which can be interpreted as a quantification of the model uncertainty and its expectation serves as the final predictive result, rather than providing a direct pointwise estimation.
Examples include prior networks~\cite{malinin2018predictive,malinin2019reverse,wu2019quantifying,nandy2020towards}, distance-aware methods~\cite{van2020uncertainty,liu2020simple}, and evidential deep learning~\cite{sensoy2018evidential,amini2020deep,deng2023uncertainty,chen2023r}.
(2) \textit{External methods}~\cite{nandy2020towards,ramalho2020density,hsu2020generalized,oberdiek2018classification,lee2020gradients} separate uncertainty quantification from the prediction task.
For example, \cite{nandy2020towards,ramalho2020density} suggest to train an additional network to estimate uncertainty on the predictions of the original network.
\cite{hsu2020generalized} obtains class wise total probabilities by applying sigmoid function to network logits, thus detecting OOD samples in the inference phase.

The previous survey~\cite{ulmer2021prior} emphasized the connection between prior networks and EDL, particularly through the Dirichlet distribution. An important property of the Dirichlet distribution is that it serves as the conjugate prior for the parameters of a Categorical distribution (or a Multinomial distribution).
This implies that when a Dirichlet prior is chosen, the posterior distribution after observing data will also be a Dirichlet distribution.
While both Prior Networks~\cite{malinin2018predictive} and EDL methods involve the construction of Dirichlet distributions, they differ significantly in their focus: Prior Networks offer tractable parameterizations of the Dirichlet prior, whereas EDL methods provide tractable parameterizations of the Dirichlet posterior.
Consequently, it is reasonable to consider EDL methods as a type of posterior network. 
In this survey, we focus on the original and intrinsic derivation form of EDL under the guidance of subjective logic theory, without delving extensively into the relationship between prior and posterior networks.

\noindent\textbf{Bayesian methods.}
As shown in \firef{related}, Bayesian uncertainty quantification methods can be mainly divided into three categories based on how to infer the generally intractable posterior distribution:
(1) \textit{Variational inference} methods~\cite{hernandez2015probabilistic,gal2016dropout,posch2019variational,nguyen2020uncertainty,dusenberry2020efficient,izmailov2021bayesian} aim to minimize the divergence between the true posterior distribution and a specified simpler, tractable distribution.
A well-known example is Monte Carlo Dropout~\cite{gal2016dropout}, which reinterprets the dropout layer as a random variable governed by a Bernoulli distribution.
By incorporating these dropout layers during both training and inference, the method provides an efficient approximation to variational inference, which effectively bridges the gap between traditional dropout techniques and Bayesian inference and thus offers a practical and scalable method for uncertainty quantification in deep neural networks.
(2) \textit{Sampling approaches}~\cite{welling2011bayesian,li2017dropout,nemeth2021stochastic} typically rely on Markov Chain Monte Carlo (MCMC)~\cite{bishop2006pattern,duane1987hybrid} techniques and provide a manner to represent the target random variable by generating samples that approximate the posterior distribution.
To obtain samples from the true posterior distribution, MCMC sampling methods generate samples iteratively in a Markov Chain manner.
(3) \textit{Laplace approximation} methods~\cite{ritter2018scalable,george2018fast,lee2020estimating,salimans2016weight,botev2017practical} simplify the posterior distribution by approximating the log-posterior around its mode using a second-order Taylor expansion, whose core is the estimation of the Hessian.
Based on the observation that most Hessian eigenvalues are frequently zero, \cite{lee2020estimating} proposes a low-rank approximation method that yields sparse representations of the covariance matrices of network layers, and illustrates that Laplace approximation can be efficiently applied to large-scale datasets. 

\noindent\textbf{Ensemble methods.} In recent years, ensemble methods have become prevalent for quantifying uncertainty in deep neural networks~\cite{lakshminarayanan2017simple,valdenegro2019deep,malinin2019ensemble,wen2020batchensemble, dusenberry2020efficient}.
The seminal work \cite{lakshminarayanan2017simple} designs member networks with dual outputs for predictions and corresponding uncertainties, demonstrating performance on par with MC Dropout~\cite{gal2016dropout} and Probabilistic Backpropagation~\cite{hernandez2015probabilistic}.
However, ensemble methods inherently demand significantly more computational resources and memory, which can be prohibitive for applications requiring rapid responses.
To address this issue, as shown in \firef{related}, several strategies have been explored:
(1)~\textit{Pruning approaches}~\cite{cavalcanti2016combining} aim to reduce the complexity of ensembles by eliminating redundant members without substantially impacting performance.
(2)~\textit{Distillation approaches}~\cite{malinin2019ensemble,lindqvist2020general} reduce the ensemble to a single model using knowledge distillation.
(3)~\textit{Sharing approaches}, such as sub-ensemble~\cite{valdenegro2019deep} and batch-ensemble~\cite{wen2020batchensemble}, aim to reduce computational and memory overhead by sharing parts of network members. 

\noindent\textbf{Post-hoc methods.} Based on the premise that diversely augmented test samples provide different perspectives and can thus capture uncertainty, post-hoc methods~\cite{shanmugam2021better,lyzhov2020greedy} typically generate multiple instances from each test sample by applying data augmentation.
These samples are then tested to compute a predictive distribution for uncertainty quantification.
Despite the evident advantage of keeping the underlying model unchanged and requiring no additional data, post-hoc methods are criticized for potentially converting many correct predictions into incorrect ones~\cite{shanmugam2021better}.

\vspace{-1mm}
\section{Theoretical Foundations of EDL}
\label{sec:foundation}
{In this section, we elaborate on the theoretical foundation of evidential deep learning (EDL), aka the subjective logic theory.
	We begin by illustrating that subjective logic extends traditional probabilistic logic by incorporating the ability to express uncertainty (\secref{subjective}), and then present the key definitions and a crucial theorem necessary for developing the uncertainty framework, subjective logic, into the uncertainty quantification method EDL in deep learning. 
	Subsequently, we discuss the relationships and differences between subjective logic and four other uncertainty reasoning frameworks (\secref{comparison}): Dempster-Shafer theory~\cite{shafer1976mathematical}, the imprecise Dirichlet model~\cite{walley1996inferences}, fuzzy logic~\cite{hajek2013metamathematics}, and Kleene's three-valued logic~\cite{fitting1994kleene}.
	Exploring these alternative uncertainty reasoning frameworks not only provides a broader context for understanding the strengths and limitations of subjective logic relative to other methods, but also encourages innovation and cross-pollination of ideas, thus paving the way for future research that integrates multiple frameworks to address complex real-world problems.}

\vspace{-1mm}
\subsection{Introduction of Subjective Logic (SL)}
\label{sec:subjective}
Drawing from Kant's philosophical concept of \textit{"das Ding an sich"} (the thing-in-itself)~\cite{hoffe2002immanuel}, we can posit that while an objective reality exists independently of us, our perception of this reality is inherently subjective, since all our knowledge and reasoning are ultimately filtered through our subjective lenses, shaped by individual experiences and cognitive limitations.
This fundamental duality between the presumed objective world and the perceived subjective world is mirrored in the various formal systems of logic and probabilistic reasoning, which attempt to bridge the gap between our subjective understanding and the objective reality by providing structured frameworks for reasoning~\cite{josang2016subjective}.

Readers may already be familiar with the \textit{probabilistic} logic in the Bayesian probabilistic theory, where an argument can take a probability value in the range $[0,1]$, thereby reflecting a degree of subjectivity by allowing the argument to be partially true.
Furthermore, subjective logic~\cite{josang2016subjective} extends probabilistic logic by incorporating not only belief and disbelief but also explicitly including uncertainty into its formalism.
Specifically, given a categorical random variable $X$ on the domain $\mathbb{X}$, a \textit{subjective opinion} in subjective logic can be formalized as an ordered triplet ${\bm \tau}=({\bm b}, u, {\bm a})$, where ${\bm b}$ is a \textit{belief mass} distribution over $X$, $u$ is a \textit{uncertainty mass}, ${\bm a}$ is a \textit{base rate} (prior distribution), and the sum of the belief mass and the uncertainty mass is limited to one.
When the domain \(\mathbb{X}\) is binary, the opinion is referred to as a binomial opinion; when the domain \(\mathbb{X}\) contains more than two elements, the opinion is termed a multinomial opinion.

Furthermore, there is a generalized form of the common \textit{binomial}/\textit{multinomial} opinions: 
Let $X$ be a hypervariable belonging to $\mathscr R(\mathbb X)=2^\mathbb X/\{X,\emptyset\}$, aka the power set of $\mathbb X$ excluding the empty set and the set $\mathbb X$ itself.
The subjective opinion 
over the variable $X$ is referred as \textit{hypernomial}. 
In this case, not only can belief mass be assigned to the whole domain $\mathbb X$ to express vacuity, it can also be assigned to other elements in the power set to express vagueness.
Besides, when the belief mass assigned to composite sets of $\mathscr R(\mathbb X)$ are zero, \ie, all belief masses are assigned to singleton classes, the hypernomial opinion is equivalent to a multinomial opinion.
Based on the concept of hypernomial subjective opinions, \cite{li2024hyper} proposes a generalized variant of EDL, named Hyper-Evidential Neural Network (HENN), which will be introduced in \secref{reformulating}.
However, in this survey, we discuss the case of \textit{multinomial} subjective opinions for simplicity without specific illustrations.

\begin{figure}
	\centering
	\includegraphics[width=\linewidth]{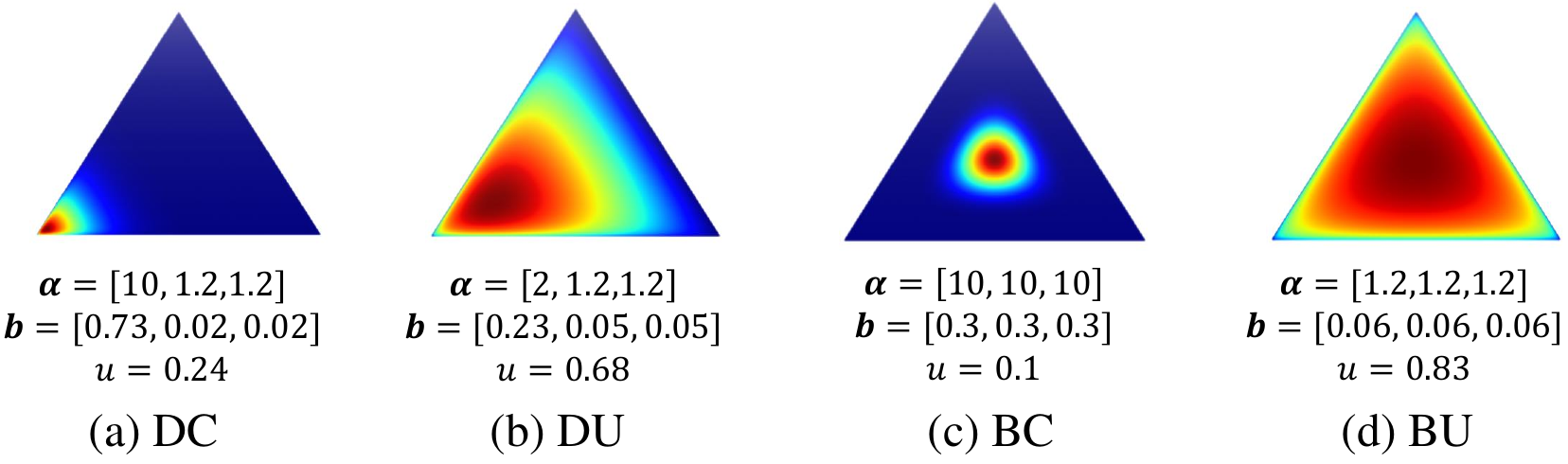}
	\caption{In 3-class classification, we present examples of Dirichlet distributions with their concentration parameters and subjective opinions across four different scenarios: (a) Dominant and Certain, (b) Dominant and Uncertain, (c) Balanced and Certain, and (d) Balanced and Uncertain.
	}
	\vspace{-2mm}
	\label{dirichlet-fig}
\end{figure}

In addition, since \textit{uncertainty mass} can be interpreted as belief mass assigned to the entire domain, subjective logic can naturally reassign the uncertainty mass $u$ into each element of domain $\mathbb X$ according to the base rate $\bm a$, resulting in a well-defined \textit{projected probability}, ${\bm P}_i = {\bm b}_i + {\bm a}_i u$, whose sum over classes is one.


With the above definition, subjective logic theory identifies a bijection between a multinomial opinion and a Dirichlet probability density function (PDF).
Formally, let $\bm p$ be a probability distribution over $\mathbb{X}$, and a Dirichlet PDF with the concentration parameter $\bm \alpha$  is denoted by $\text{Dir}({\bm p, \bm \alpha})$:  
\begin{equation}
	\label{dirichlet}
	\text{Dir}(\bm p,\bm \alpha)=
	\frac{\Gamma\left(\sum_{i\in\mathbb{X}}\bm \alpha_i\right)}
	{\prod_{i\in\mathbb{X}}\Gamma(\bm \alpha_i)}
	\prod_{i\in\mathbb{X}}\bm p_i^{\bm \alpha_i-1},
\end{equation}
where $\Gamma$ denotes the Gamma function, $\bm \alpha_i \geq 0$, and $\bm p_i\neq0$ if $\bm \alpha_i <1$.
Then, given the base rate $\bm a$, there exists a bijection $F$ between the opinion $\bm \tau$ and the Dirichlet PDF $\text{Dir}({\bm p, \bm \alpha})$,
where $\bm \alpha$ satisfies $\bm \alpha_i = \bm b_i W/u + \bm a_iW$, and $W$ is a positive prior weight.
This relationship arises from the interpretation of second-order uncertainty through probability density, and is crucial in the framework of subjective logic because it facilitates calculus reasoning using PDFs.
Examples of Dirichlet distributions, along with their respective concentration parameters and subjective opinions across various scenarios in 3-class classification (\eg, classification results dominated by a single category or evenly distributed across categories, and whether the predictions are certain or uncertain), are illustrated using triangle heatmaps in \firef{dirichlet-fig}.
Specifically, the vertices of the heatmaps correspond to different categories, and each point within the heatmap represents a specific allocation of class probabilities along with its corresponding PDF value from the Dirichlet distribution.


\vspace{-1mm}
\subsection{Other Uncertainty Reasoning Frameworks}
\label{sec:comparison}
\textbf{Comparison with Dempster-Shafer Theory (DST)~\cite{shafer1976mathematical}.} The DST, often referred to as evidence theory, was initially introduced by Dempster within the realm of statistical inference~\cite{dempster2008upper}, and Shafer later expanded this theory into a comprehensive framework for representing epistemic uncertainty~\cite{shafer1976mathematical}.
DST has been pivotal in shaping subjective logic by challenging the traditional additivity principle of probability theory.
Specifically, DST allows the sum of probabilities for all mutually exclusive events to be less than one.
This feature enables both DST and subjective logic to explicitly represent uncertainty about probabilities by allocating belief mass to the entire domain.
The difference between DST and subjective logic is that, subjective logic encourages the evidence distribution of samples with high uncertainty to fall back onto a prior, while DST does not include a flexible base rate representing the prior distribution.

\noindent\textbf{Comparison with Imprecise Dirichlet Model (IDM)~\cite{walley1996inferences}.}  The IDM for multinomial variables derives upper and lower probabilities by adjusting the minimum and maximum base rates in the Beta/Dirichlet PDF for each possible value within the domain.
Unlike subjective logic, which employs a prior weight to influence the base rate's effect, IDM creates an interval of expected probabilities by setting the base rate to its maximum (equal to one) for upper probabilities, and to zero for lower probabilities.
Note that the intervals provided by IDM are not strictly bounded, meaning the actual probabilities may fall outside these estimated ranges.

\noindent\textbf{Comparison with Fuzzy Logic~\cite{hajek2013metamathematics}.} In Fuzzy Logic, variables are defined by terms that have imprecise and partially overlapping meanings. For instance, when considering the variable temperature, potential values might include "Low (0 to \qty{20}{\degreeCelsius})", "Medium (15 to \qty{30}{\degreeCelsius})", and "High (25 to \qty{40}{\degreeCelsius})".
Despite the inherent fuzziness of these values, temperature can still be represented in an exact and crisp manner using a \textit{fuzzy membership function}. For example, one could state "The temperature is 0.3 Low and 0.7 Medium", which quantitatively expresses the degree to which the temperature belongs to each vague category.
Conversely, in subjective logic, values are inherently crisp, but subjective opinions incorporate an uncertainty mass to capture ambiguity.
Fuzzy logic and subjective logic address different aspects of uncertainty, and there is potential to integrate these two approaches by representing fuzzy membership functions using subjective opinions~\cite{josang2008continuous}.

\noindent\textbf{Comparison with Kleene’s Three-Valued Logic~\cite{fitting1994kleene}.}
In this theory, 
propositions are categorized as either TRUE, FALSE, or UNKNOWN.
A significant limitation of this system is that it broadly labels all non-absolute propositions as UNKNOWN, without providing a detailed quantification of the degree of uncertainty,
which leads to a clear issue when dealing with the conjunction of numerous UNKNOWN propositions.
Subjective logic addresses this paradox effectively. When a series of vacuous opinions are combined, the resulting base rate diminishes towards zero, which in turn minimizes the projected probability.

\vspace{-1mm}
\section{Theoretical Explorations of EDL}
\label{sec:the_exp}
\begin{figure*}[thbp]
	\centering
	\includegraphics[width=0.8\textwidth]{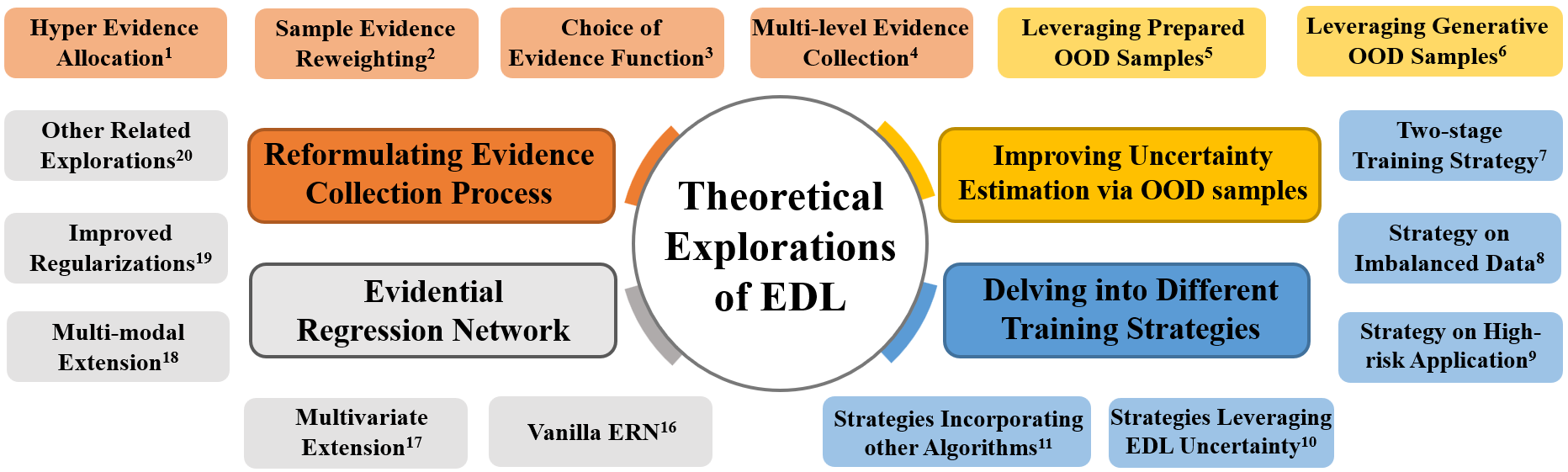}
	\vspace{-1mm}
	\caption{Structure of theoretical explorations of evidential deep learning, including:
		(1) Reformulating evidence collection process ($^1$\cite{li2024hyper}, $^2$\cite{deng2023uncertainty}, $^3$\cite{pandey2023learn}, $^4$\cite{shen2023post,gao2023vectorized});
		(2) Improving uncertainty estimation via OOD samples ($^5$\cite{nagahama2023learning,zhao2019quantifying}, $^6$\cite{sensoy2020uncertainty,hu2021multidimensional,davies2023knowledge}); 
		(3) Delving into different training strategies ($^7$\cite{li2022tedl}, $^8$\cite{xia2022hybrid}, $^9$\cite{sensoy2021misclassification}, $^{10}$\cite{pandey2022multidimensional,chen2022dual},$^{11}$\cite{kandemir2022evidential,haussmann2019bayesian});
		(4) Evidential regression network ($^{12}$\cite{amini2020deep}, $^{13}$\cite{meinert2021multivariate}, $^{14}$\cite{ma2021trustworthy}, $^{15}$\cite{ye2024uncertainty,wu2024evidence,oh2022improving}, $^{16}$\cite{meinert2023unreasonable,duan2024evidential,pandey2023evidential}).
	}
	\label{theory}
	\vspace{-3mm}
\end{figure*}

In this section, we provide a comprehensive overview of the theoretical advancements in evidential deep learning (EDL) since its inception by Sensoy et al. \cite{sensoy2018evidential}.
We begin with an introduction to the model construction, model optimization, and model simplification of the EDL method (\secref{evidential}).
Subsequently, as shown in \firef{theory}, we categorize the theoretical explorations of EDL into five main categories based on their unique characteristics in various aspects:
(1) evidence collection, if the evidence collection process has been reformulated (\secref{reformulating});
(2) evidence source, if additional out-of-distribution (OOD) samples have been leveraged to improve uncertainty estimation (\secref{improving});
(3) evidence use, if different training strategies have been adopted or designed (\secref{delving});
and (4) regression task, if the explorations centered around regression tasks as opposed to traditional classification tasks (\secref{regression}).

\vspace{-1mm}
\subsection{Evidential Deep Learning}
\label{sec:evidential}
Evidential Deep Learning (EDL) explores the direct application of subjective logic theory to deep neural networks.
Specifically, \cite{sensoy2018evidential}, which we refer to as vanilla EDL, trains a neural network to function as an analyst, capable of producing reliable belief mass $\bm{b}$ and uncertainty mass $u$ for test samples.
For instance, in a classification task involving $C$ classes, given an input sample $\bm{x}$, the network can provide the evidence $\bm e=[\bm e_1,...,\bm e_C]\in\mathbb R^C_+$, where $\bm e_i$ represents the amount of evidence supporting the claim that \textit{"the sample $\bm x$ belongs to the $i$-th category"}, \eg, $\bm e=\text{Softplus}(f(\bm x))$, where $f$ is the deep neural network, Softplus is an activation function, sometimes also termed the evidence function, which ensures the non-negative property of evidence and can be replaced by other non-negative activation functions like ReLU.
Note that \textit{evidence} in EDL has no relevance with \textit{model evidence} in Bayesian theory, which denotes the marginal likelihood.
Actually, \textit{observation} may be a more appropriate term to avoid ambiguity, since in EDL $\bm e_i$ can be interpreted as the number of observations of the event that \textit{"a random variable takes the value $i$"}, $i\in \mathbb X=[1,...,C]$.

Consequently, the belief mass for classifying $\bm{x}$ into the $i$-th class, as well as the uncertainty mass, which indicates the extent to which the model is uncertain about the category of $\bm{x}$, can be derived from evidence $\bm e$ as follows:
\begin{equation}
	\label{edl-mass}
	\bm b_i = \frac{\bm e_i}{\sum_{j\in \mathbb X} \bm e_j + W},\quad u= \frac{W}{\sum_{j\in \mathbb X} \bm e_j + W},
\end{equation}
where $W$ is a positive scalar representing the prior weight.

As introduced in \secref{subjective}, there exists a bijection between the Dirichlet PDF denoted $\text{Dir}(\bm p, \bm \alpha)$ and the opinion $\bm \tau=(\bm b, u, \bm a)$.
Specifically, equipped with \eqnref{edl-mass}, we can derive the relationship between the parameter vector of the Dirichlet PDF and the EDL evidence as $\bm \alpha_i = \bm e_i + \bm a_i W$.
Moreover, since~\cite{sensoy2018evidential} sets the base rate $\bm a_i$ as a uniform distribution over the domain $\mathbb X$, aka $\bm a_i = 1/C$, and sets the prior weight $W$ as the cardinality of domain $\mathbb X$, aka the class number $C$, the above relationship can be simplified into the most common form in EDL-related literature:
\begin{equation}
	\label{addone}
	\bm \alpha_i = \bm e_i + 1,\quad\forall i\in \mathbb X.
\end{equation}

To obtain the loss function for model optimization, the vanilla EDL method integrates traditional loss functions over the class probability $\bm p$ which follows the above Dirichlet distribution.
\cite{sensoy2018evidential} highlights that the following EDL loss formulation $\mathcal{L}_{\text{mse-edl}}$, which is obtained by integrating the traditional mean square error (MSE) loss over the Dirichlet distribution, generally yields satisfactory results:
\begin{equation}
	\label{edl-loss}
	\begin{aligned}
		\mathcal{L}_{\text{mse-edl}}
		&=\frac{1}{|\mathcal{D}|}\sum_{(\bm x, \bm y)\in \mathcal{D}}\mathbb{E}_{\bm p\sim\text{Dir}(\bm p, \bm \alpha)}\left[\Vert \bm y - \bm p\Vert_2^2\right] \\
		&=\frac{1}{|\mathcal{D}|}\sum_{(\bm x, \bm y)\in \mathcal{D}}\sum_{i\in\mathbb X}\left(\bm y_i -\frac{\bm \alpha_i}{S}\right)^2 + \frac{\bm \alpha_i(S - \bm \alpha_i)}{\bm \alpha_i^2(\bm \alpha_i+1)},
	\end{aligned}
\end{equation}
where the training set $\mathcal{D}$ consists of sample features and their one-hot labels denoted $(\bm x, \bm y)$, and $S$ is the sum of $\bm \alpha_i$ over $i\in\mathbb X$.
Although empirical results tend to favor the MSE loss, other formulations of EDL loss functions have also been investigated. Specifically, integrating the cross-entropy (CE) loss over the Dirichlet distribution results in:
\begin{equation}
	\label{edl-loss-ce}
	\begin{aligned}
	\mathcal{L}_{\text{ce-edl}}
	&=\frac{1}{|\mathcal{D}|}\sum_{(\bm x, \bm y)\in \mathcal{D}}\sum_{i\in\mathbb X}\bm y_i \left(\psi(S) - \psi(\bm \alpha_i)\right),
	\end{aligned}
\end{equation}
where $\psi(\cdot)$ is the digamma function.
Besides, the negative likelihood loss in the EDL framework can be calculated by integrating class probabilities over the Dirichlet  distribution: 
\begin{equation}
	\label{edl-loss-nll}
	\begin{aligned}
	\mathcal{L}_{\text{nll-edl}}
	&=\frac{1}{|\mathcal{D}|}\sum_{(\bm x, \bm y)\in \mathcal{D}}\sum_{i\in\mathbb X}\bm y_i \left(\log(S) - \log(\bm \alpha_i)\right).
	\end{aligned}
\end{equation}

Finally, EDL-related works commonly adopts an additional regularization term $\mathcal{L}_{\text{kl}}$ to suppress the evidence of non-target classes by minimizing the Kullback-Leibler (KL) divergence between a modified Dirichlet distribution parameterized by $\tilde{\bm \alpha}_X=\bm y + (\bm 1 - \bm y)\odot \bm \alpha_X$, {where $\odot$ represents the Hadamard product,} and a uniform distribution.
The formulation of $\tilde{\bm \alpha}_X$ indicates that the parameter of the target class has been set to 1 and others are left unchanged.
Specifically, the regularization term has the following form:
\begin{equation}
	\label{kl-divergence}
	\begin{aligned}
		\mathcal{L}_{\text{kl}}
		&=\frac{1}{|\mathcal{D}|}\sum_{(\bm x, \bm y)\in \mathcal{D}}\text{KL}\left(\text{Dir}(\bm p, \tilde{\bm \alpha}), \text{Dir}(\bm p, \bm 1)\right). \\
	\end{aligned}
\end{equation}
Therefore, the optimization objective of vanilla EDL is set as $\mathcal{L}_\text{edl} + \mu_t\mathcal{L}_\text{kl}$, where $\mu_t=\min(1.0, t/10)\in[0,1]$ is the annealing coefficient, $t$ is the training epoch index.

In inference, EDL only requires a single forward pass to calculate the uncertainty.
Specifically, the projected probability~$\bm P$ defined in Definition~2 is adopted as the predictive scores, and \eqnref{edl-mass} is used to calculate the uncertainty mass $u$ as the uncertainty of classification,
\begin{equation}
	\label{edl-probability}
	\bm P_i=\frac{\bm e_i + 1}{\sum_{j\in \mathbb X}\bm e_j + C}=\frac{\bm \alpha_i}{S},\quad u= \frac{C}{\sum_{i\in \mathbb X}\bm e_i + C}=\frac{C}{S}.
\end{equation}

Furthermore, \cite{chen2023r} proposes a generalized version named Relaxed-EDL (R-EDL), which relaxes two non-essential settings of the traditional EDL method in the model construction and optimization stages.
On one hand,
it is observed that the scalar $1$ added to the evidence arises from a rigid operation of setting the prior weight $W$ equal to the cardinality of the domain $C$, which is not mandated by the subjective logic theory and may result in unreasonable results.
Given the myriad of complex factors influencing the network's output, \cite{chen2023r} abandons the rigid operation
but instead to treat $W$ as an adjustable hyper-parameter within the neural network.
On the other hand, the variance-minimized regularization term in \eqnref{edl-loss} is also deprecated to alleviate over-confidence.
Formally, the optimization objective of R-EDL is simply given by:
\begin{equation}
	\label{redl-loss}
	\mathcal{L}_{\text{redl}}
	=\frac{1}{|\mathcal{D}|}\sum_{(\bm x, \bm y)\in \mathcal{D}}\sum_{i\in\mathbb X} \left(\bm y_i - \bm P_i\right)^2.
\end{equation}
It is demonstrated that both relaxations are effective in achieving more precise uncertainty quantification.

\vspace{-2mm}
\subsection{Reformulating Evidence Collection Process}
\label{sec:reformulating}

As a core step in EDL algorithms, the evidence collection process significantly impacts the quality of the uncertainty in the model's output. Beyond the traditional evidence collection methods discussed in Section 3.1, existing research has conducted in-depth explorations into various aspects, such as hyper evidence 
allocation~\cite{li2024hyper}, sample evidence reweighting~\cite{deng2023uncertainty}, the choice of evidence functions~\cite{pandey2023learn}, and multi-level evidence collection~\cite{shen2023post}. In the following sections, we will provide a detailed overview of these studies.

Based on hypernomial subjective opinions, introduced as Definition 2 in \secref{subjective}, \cite{li2024hyper} proposes a generalized variant of EDL, named the Hyper-Evidential Neural Network (HENN).
The extension from \textit{"multinomial"} to \textit{"hypernomial"} is straightforward: it only involves replacing the domain $\mathbb{X}$ with its reduced powerset $\mathscr{R}(\mathbb{X}) = 2^\mathbb{X} / \{\mathbb{X}, \emptyset\}$, which is the power set of $\mathbb{X}$ excluding the empty set and the set $\mathbb{X}$ itself.
When the belief mass assigned to composite sets of $\mathscr R(\mathbb X)$ are zero, \ie, all belief masses are assigned to singleton classes, the hypernomial opinion is equivalent to a multinomial opinion.
With this extension, not only can belief mass be assigned to the whole domain $\mathbb X$ to express vacuity, it can also be assigned to other elements in the reduced power set to express vagueness.
The bijective hypernomial Dirichlet distribution has the following form:
\begin{equation}
	\text{HyperDir}(\bm p,\bm \alpha)=Z_h^{-1}
	\prod_{i\in\mathscr R(\mathbb X)}\bm p_i^{\bm \alpha_i-1},
\end{equation}
where $Z_h$ is the normalization constant.

HENN~\cite{li2024hyper} explores a simple case of the hypernomial subjective opinions, where the composite sets $\{\mathcal S_1,...,\mathcal S_\eta\}$ in $\mathscr R(\mathbb X)$ represent a partition of singleton classes, \ie, $\cup_{j=1}^\eta \mathcal S_j=\mathbb X$ and $\mathcal S_i \cap \mathcal S_j=\emptyset$.
In this case, the multinomial Dirichlet distribution in the vanilla EDL method will be transformed into a special hypernomial Dirichlet distribution, termed as a grouped Dirichlet distribution (GDD), whose PDF has the following form:
\begin{equation}
	\text{GDD}(\bm p, \bm \alpha, \bm c) = Z^{-1} \prod_{i\in\mathbb X} \bm p_i^{\bm \alpha_i - 1} \prod_{j=1}^{\eta} \left( \sum_{l \in S_j} \bm p_l \right)^{\bm c_j},
\end{equation}
where $\bm c_j\in\mathbb R^+$ is the concentration parameter of composite set $\mathcal S_j$, $Z = \left[ \prod_{j=1}^{\eta} B \left( \{ \alpha_l \}_{l \in S_j} \right) \right] B \left( \{ \beta_j \}_{j=1}^{\eta} \right)$ is the normalization constant, where $\beta_j = \sum_{l \in S_j} \alpha_l + c_j$, and $B(\cdot)$ is the beta function.
Given the binary vector representation~$\tilde{\bm y}\in\{0,1\}^K$ of multi-class labels, the optimization objective, termed as uncertainty partial cross-entropy (UPCE), is formulated by integrating the partial cross-entropy loss~\cite{cour2011learning} over the above GDD distribution:
\begin{equation}
	\begin{aligned}
		&\mathcal{L}_\text{UPCE}(\bm x, \tilde{\bm y}) 
		= \mathbb{E}_{\bm p \sim \text{GDD}(\bm{p}, \bm{\alpha}, \bm{c})} \left[ -\sum_{i\in\mathbb X} \tilde{y}_i \log p_i \right]. \\
	\end{aligned}
\end{equation}

$\mathcal{I}$-EDL proposed by \cite{deng2023uncertainty} modifies the vanilla EDL method by incorporating Fisher information matrix (FIM) to measure the informativeness of evidence carried by samples.
Formally, the FIM is defined as:
\begin{equation}
	\mathcal{I}(\bm \alpha)
	= \mathbb{E}_{\text{Dir}(\bm p|\bm \alpha)}\left[\frac{\partial l}{\partial \bm \alpha}\frac{\partial l}{\partial \bm \alpha^T}\right],
\end{equation}
where $l = \log \text{Dir}(\bm p|\bm \alpha)$ is the log-likelihood.
Based on the motivation that class labels with higher evidence should be assigned with large variances, the inverse of the FIM is adopted as the variance of the distribution of $\bm y$, that is, the target variable $\bm y$ is assumed to follow a multivariate Gaussian distribution: $\bm y\sim\mathcal{N}(\bm p,\sigma^2\mathcal{I}(\bm \alpha)^{-1})$.
In this way, the optimization objective $\mathcal{L}_{\mathcal{I}-\text{edl}}$ can be derived as:
\begin{equation}
	\small
	\label{iedl}
	\begin{aligned}
	&\frac{1}{|\mathcal{D}|}\sum_{(\bm x, \bm y)\in \mathcal{D}}\sum_{i\in\mathbb X}\left(\left(\bm y_i -\frac{\bm \alpha_i}{S}\right)^2 
	+ \frac{\bm \alpha_i(S - \bm \alpha_i)}{\bm \alpha_i^2(\bm \alpha_i+1)}\right)\psi^{(1)}(\bm \alpha_i) \\
	&-\lambda_1 \mathcal{L}_{\mathcal{I}}
	+\lambda_2 \mathcal{L}_{\text{kl}},
	\end{aligned}
\end{equation}
where $\psi^{(1)}(x)=d^2\ln\Gamma(x)/dx^2$ denotes the trigamma function, $\lambda_1$ and $\lambda_2$ are hyper-parameters, $\mathcal{L}_{\text{kl}}$ is consistent with \eqnref{kl-divergence}, and $\mathcal{L}_{\mathcal{I}}
= \frac{1}{|\mathcal{D}|}\sum_{(\bm x, \bm y)\in \mathcal{D}}\sum_{i\in\mathbb X}\log|\mathcal{I}(\bm \alpha_i)|$.
In \eqnref{iedl}, the FIM-based term $\psi^{(1)}(\bm \alpha+i)$ acts as an adaptive weight, encouraging the predictive probability of a class with low evidence to be more accurate.
The penalty term $\mathcal{L}_{\mathcal{I}}$ serves as a regularization term to prevent excessive evidence that could lead to overconfidence.
Experiments demonstrate that $\mathcal{I}$-EDL outperforms the traditional EDL method on various metrics of uncertainty estimation, particularly in the few-shot classification setting.

\cite{pandey2023learn} observes that evidence functions generating zero evidence regions can inhibit evidential neural networks from extracting useful information from training samples within these regions.
The study focuses on three activation functions: ReLU, Softplus, and Exp, which demonstrate progressively better performance.
ReLU activation is less effective as it completely eliminates all information from negative logits and creates the largest zero-evidence region in the evidence space, thereby leaving training data with no evidence in this area.
Softplus activation reduces the size of the zero-evidence region and thus performs better than ReLU; however, it may struggle to correct acquired knowledge when the model has strong erroneous evidence.
Exponential activation has the smallest zero-evidence region and avoids the aforementioned issues, leading to the best performance. Consequently, most recent EDL-related research favors Softplus and Exp over the ReLU function used by \cite{sensoy2018evidential} as the evidence functions.

Since different intermediate layers of a base model capture varying levels of feature representations, leveraging this diversity is essential for comprehensive evidence collection and uncertainty quantification. 
Inspired by this observation, \cite{shen2023post} suggests gathering evidence from multiple intermediate layers instead of solely from the final network layer.
The proposed model includes $m$ linear layers, denoted as $\bm g_j$, $j=1,...,m$, connected to different intermediate layers of the base model, and a final linear layer $\bm g_c$ which collects evidence from the  combination of all the above features.
Specifically, given an input sample $\bm x$ and multiple intermediate features $\{ \Phi_j (\bm x) \}_{j=1}^m$ extracted from the base model, the final output evidence can be formulated as:
\begin{equation}
	\bm e
	= \bm g(\{\Phi_j (\bm x)\}_{j=1}^m)
	= \bm g_c (\{\bm g_j (\Phi_j (\bm x)) \}_{j=1}^m).
\end{equation}
Note that the linear layers $\bm g_j$ and $\bm g_c$ only consist of fully connected layers and activation functions, ensuring a simpler model structure and enabling efficient training.
The optimization objective is the cross-entropy loss (\eqnref{edl-loss-ce}) of the EDL form accompanied by a regularization term to prevent overconfidence.
Similarly, rather than merely leveraging the output logits of the last layer as an evidence source, \cite{gao2023vectorized} proposes a vectorized EDL version and designs a series of learnable meta units that serve as fundamental elements constituting diverse categories.
Subsequently, a local-to-global evidence collection approach is proposed to perform uncertainty estimation. 

\vspace{-1mm}
\subsection{ Improve Uncertainty Estimation via OOD samples}
\label{sec:improving}

In the ideal case, an uncertainty estimator should assign higher uncertainty to both difficult samples and out-of-distribution (OOD) samples, while assigning lower uncertainty to easily classified samples and in-distribution (ID) samples.
Unfortunately, \cite{sensoy2020uncertainty} observes that although the vanilla EDL method effectively decreases prediction confidence when classifying difficult samples near the class boundary, it still maintains high prediction confidence when tested with OOD samples.
To address these issues, existing works~\cite{nagahama2023learning,sensoy2020uncertainty,hu2021multidimensional,zhao2019quantifying,davies2023knowledge} have explored various methods to improve uncertainty estimation by incorporating prepared or generative OOD samples into the training process.

When training data includes prepared OOD samples, \cite{nagahama2023learning} presents a simple m-EDL paradigm that manually adds an additional class to leverage these samples. The modified optimization objective is formulated as:
\begin{equation}
	\begin{aligned}
	\mathcal{L}_{\text{m-edl}}
	=&\frac{1}{|\mathcal{D}|}\sum_{(\bm x, \bm y)\in \mathcal{D}}\left(\sum_{i\in\mathbb X}\bm y_i \left(\log(S) - \log(\bm \alpha_i)\right)\right)\\
	&+ \bm y_u(\log(S) - \log(\bm \alpha_u)),
	\end{aligned}
\end{equation}
where $S = \sum_i\bm\alpha_i$, $\bm \alpha_u = K+1$, and $\bm y_u\in\{0,1\}$ is the ground-truth label indicating whether the input sample belongs to unknown categories.
In fact, m-EDL incorporates a regularization term to suppress evidence collection on OOD samples.
Similarly, \cite{zhao2019quantifying} proposes a regularized ENN (evidential neural network) method that encourages the model to assign high vacuity to OOD samples and high dissonance to samples near the classification boundary.
The design of the loss function is straightforward:
\begin{equation}
	\mathcal{L}_{\text{mse-edl}} + \lambda_1\mathbb{E}_{\bm x \sim \mathcal{D}_{\text{OOD}}}[\text{Vac}(\bm x)] + \lambda_2\mathbb{E}_{\bm x \sim \mathcal{D}_{\text{BOD}}}[\text{Diss}(\bm x)],
\end{equation}
where $\mathcal{D}_{\text{OOD}}$ denotes the set of OOD training samples, $\mathcal{D}_{\text{BOD}}$ represents the set of samples with conflicting evidence, and $\lambda_{1/2}$ are trade-off hyper-parameters.
$\text{Vac}(\bm x)$ represents the vacuity of sample $\bm x$, aka the uncertainty mass, while $\text{Diss}(\bm x)$ refers to the dissonance of $\bm x$, measuring the extent of evidence contradiction. They can be calculated as follows:
\begin{equation}
	\small
	\label{vac}
	\text{Vac}(\bm x) = \frac{W}{S},\quad\text{Diss}(\bm x)=\sum_{i\in\mathbb{X}}\left(\frac{\bm b_i\sum_{j\in\mathbb X\backslash i}\bm b_j\text{Bal}(j,i)}{\sum_{j\in\mathbb X\backslash i}\bm b_j}\right),
\end{equation}
where $\text{Bal}(j,i)$ is referred to as the relative mass balance between belief masses $\bm b_j$ and $\bm b_i$.
The expressions of vacuity and dissonance in \eqnref{vac} are also adopted by other works. For example, \cite{pandey2022multidimensional} utilizes the convex combination of $\text{Vac}(\bm x)$ and $\text{Diss}(\bm x)$ to estimate the total uncertainty for task selection in meta-learning:
\begin{equation}
	\text{Uct}(\bm x) = \lambda\cdot \text{Vac}(\bm x) + (1 - \lambda)\cdot\text{Diss}(\bm x).
\end{equation}
$\lambda$ is a trade-off hyper-parameter that is set to a relatively high value during the early phase of meta-learning to better explore the task space.
As training progresses, $\lambda$ decreases to emphasize more difficult tasks.

Although effective, the above methods~\cite{nagahama2023learning,zhao2019quantifying} have the drawback of requiring existing OOD samples for training, which are not always available in real-world applications.
To address this issue, \cite{sensoy2020uncertainty} utilizes the latent space of a variational autoencoder (VAE)~\cite{doersch2016tutorial} as a surrogate for semantic similarity between samples in the input space, thus generating OOD samples that are similar to, yet distinctly separable from, the training samples.
The training process incorporates a generator for OOD samples and a discriminator that attempts to distinguish real samples from the generated ones, akin to the framework of generative adversarial networks (GAN)\cite{goodfellow2020generative}.
Unlike \cite{sensoy2020uncertainty} which only leverages close OOD samples, \cite{hu2021multidimensional} argues that both close and far-away samples are equally important to achieve comprehensive uncertainty estimation.
Specifically, \cite{hu2021multidimensional} proposes a regularized EDL paradigm, whose loss function over the model parameters $\Theta$ of the model's function $f$ is given by:
\begin{equation}
	\small
	\label{OOD1}
	\mathbb{E}_{\bm x, \bm y \sim \mathcal{D}_{\text{in}}} \left[ \mathcal{L}_{\text{edl}}(\bm x, \bm y \mid f, \Theta) \right] 
	- \beta \mathbb{E}_{\hat{\bm x} \sim \mathcal{D}_{\text{out}}} \left[\mathrm{Vac}(f(\hat{\bm x} \mid \Theta)) \right],
\end{equation}
where $\mathcal{D}_{\text{in/out}}$ represents the set of ID/OOD data, $\text{Vac}$ denotes the uncertainty mass, and $\beta$ is a trade-off hyper-parameter.
It is evident that, in addition to the traditional EDL loss function $\mathcal{L}_{edl}$, \eqnref{OOD1} includes an additional regularization term designed to produce high uncertainty for OOD samples.
To provide sufficient various OOD samples, \cite{hu2021multidimensional} adopts a Wasserstain generative adversarial network (WGAN)~\cite{arjovsky2017wasserstein}, whose optimization objective with an uncertainty regularization is formulated as:
\begin{equation}
	\label{OOD2}
	\begin{aligned}
	\min_{G} \max_{D} \mathbb{E}_{\bm x \sim \mathcal D_\text{in}} \left[ D(\bm x) \right] 
	- \mathbb{E}_{\hat{\bm x} \sim \mathcal D_\text{gen}} \left[ D(\hat{\bm x}) \right]  \\
	- \beta \mathbb{E}_{\hat{\bm x} \sim \mathcal D_\text{gen}} \left[ \mathrm{Vac}(f(\hat{\bm x} \mid \Theta)) \right],
	\end{aligned}
\end{equation}
where $D$ and $G$ represent the discriminator and generator components of the WGAN, respectively, and $\mathcal D_\text{gen}$ represents the generated OOD data. To ensure that $G$ recovers $\mathcal D_\text{out}$, an uncertainty regularization term is also included.
The regularized EDL loss (\eqnref{OOD1}) and the regularized WGAN loss (\eqnref{OOD2}) are jointly trained, which enables the evidential neural network to effectively leverage the diverse types of OOD samples generated by the WGAN.

\vspace{-1mm}
\subsection{Delving into Different Training Strategies}
\label{sec:delving}
Beyond the methods and sources of evidence collection, model training strategies also play a crucial role in influencing the performance of EDL methods~\cite{li2022tedl}.
In this section, we review several studies that have explored various training strategies to enhance uncertainty estimation quality under specific conditions, such as imbalanced data~\cite{xia2022hybrid} and high-risk applications~\cite{sensoy2021misclassification}. 
Additionally, it is noteworthy that the estimated uncertainty in EDL serves as a natural metric for assessing the difficulty levels of training samples.
This characteristic can be leveraged to design diverse training strategies, providing significant benefits~\cite{pandey2022multidimensional,chen2023r}.
Moreover, some studies have sought to integrate existing machine learning algorithms into the EDL paradigm to design new training strategies~\cite{kandemir2022evidential,haussmann2019bayesian}.

To harness the potential of training, \cite{li2022tedl} introduces a two-stage training paradigm for evidential deep learning (TEDL), wherein the first stage aims to achieve accurate point estimates by utilizing the traditional cross-entropy loss, and the second stage focuses on quantifying uncertainty using a reformulated EDL loss that incorporates the Exponential Linear Unit (ELU)~\cite{clevert2015fast} as the evidence function.
This approach is intuitively reasonable: using the traditional optimization objective in the early stages of training allows the model to reach a stable state with high accuracy. Building upon this foundation, switching to the EDL optimization objective enables the model to leverage the magnitude information of the output logits, thereby achieving uncertainty estimation.
Through experiments conducted on binary classification tasks, it is demonstrated that TEDL achieves superior  accuracy compared to standard EDL and exhibits improved robustness to variations in hyper-parameters.

Aiming to achieve less biased categorical predictions on imbalanced data, \cite{xia2022hybrid} introduces the Hybrid-EDL method, which integrates training-phase data augmentation and validation-phase calibration into the standard EDL approach. 
Specifically, the method balances class frequencies in the training set by randomly reusing samples from minority classes. 
Additionally, the classification evidence for minority classes are post-hoc calibrated by evaluating the class-wise performance on the validation set.

By incorporating learnable prior counts and misclassification risks into the vanilla EDL method, \cite{sensoy2021misclassification} proposes a loss function for training evidential classifiers.
On one hand, \eqnref{addone} which formulates the parameters of the constructed Dirichlet distribution, is reformulated as:
\begin{equation}
	\label{OOD3}
	\bm \alpha(\bm x) = \bm e(\bm x) + \bm \gamma(\bm x) = \bm e(\bm x) + K\text{softmax}(\bm Wf^\prime(\bm x) + \bm b),
\end{equation}
where $\bm W$ and $\bm b$ are additional weight and bias variables, and $f^\prime(\bm x)$ is the input to the logit layer.
In this manner, the product of the prior weight and the uniform prior distribution in the vanilla EDL method, which is always $\bm 1$, is redistributed per sample across all potential categories as $\bm \gamma$.
This reformulation shares a similar motivation with \cite{chen2023r}, which also relaxes the constraints 
by setting the prior weight as a hyper-parameter.
On the other hand, integrating the misclassification risks over the Dirichlet distribution parameterized by $\bm \alpha_i$ as given by \eqnref{OOD3}, we can obtain the expected risk as:
\begin{equation}
	\mathbb E[\text{risk}(\bm x)] = \frac{\sum_{i\in\mathbb X}R_{yi}(\bm e(\bm x) + \bm \gamma(\bm x))}{K + \sum_{i\in\mathbb X}\bm e(x)},
\end{equation}
where $R_{yi}$ represents the misclassification risk or cost of misclassifying a sample from category $y$ to category $i$.
Therefore, the final loss function is given by:
\begin{equation}
	\small
	\begin{aligned}
	&\mathcal{L}_{\text{risk}} 
	= \frac{1}{|\mathcal{D}|}\sum_{(\bm x, \bm y)\in \mathcal{D}}\sum_{i\in\mathbb X}\bm y_i \left(\psi(S) - \psi(\bm \alpha_i)\right) + \\
	& \text{KL}\left(\text{Dir}(\bm p, \bm \alpha), \text{Dir}(\bm p, \bm e\odot\bm y + \bm 1)\right)
	+ \kappa \sum_{i\in\mathbb X}R_{yi}(\bm e + \bm \gamma),
	\end{aligned}
\end{equation}
which optimizes the evidential classifier while minimizing the misclassification risk.
Note that {$\odot$ represents the Hadamard product}, $\kappa$ is a trade-off hyper-parameter for risk regularization, and the denominator of the expected risk is omitted to prevent the model from generating excessive evidence for less risky categories when it outputs incorrect predictions.
Although the KL-divergence regularization differs from the form in \eqnref{kl-divergence}, it similarly aims to suppress the evidence for non-target categories.

Additionally, prior studies have investigated the role of estimated uncertainty in EDL for designing training strategies.
For instance, with a dynamic hyper-parameter $\lambda$, \cite{pandey2022multidimensional} defines total uncertainty $\text{Uct}(\bm x)$ for task selection in meta-learning as a convex combination of $\text{Vac}(\bm x)$ and $\text{Diss}(\bm x)$, specifically $\text{Uct}(\bm x) = \lambda \cdot \text{Vac}(\bm x) + (1 - \lambda) \cdot \text{Diss}(\bm x)$.
Initially, $\lambda$ is set to a high value to facilitate extensive exploration of the task space.
Over time, $\lambda$ is reduced to concentrate on more challenging tasks, aiding in model fine-tuning.
Similarly, in the context of weakly-supervised learning, \cite{chen2022dual} leverages the fine-grained uncertainty order to sequentially focus on entire samples, achieving progressive learning.

Moreover, some studies have sought to integrate existing machine learning algorithms into the EDL paradigm to design new training strategies.
For example, \cite{kandemir2022evidential} extends EDL by incorporating neural processes and neural Turing machines, thus proposing the Evidential Tuning Process which shows strong performances but requires a rather complex memory mechanism.
\cite{haussmann2019bayesian} advances EDL by incorporating a Bayesian neural network framework, termed Bayesian Evidential Deep Learning (BEDL), aiming for enhanced accuracy and improved uncertainty quantification.
In particular, BEDL introduces a local prior on the EDL weights $\bm w_n$ with shared hyper-parameters, and optimizes the marginal likelihood, essentially employing a Bayesian local neural net on the likelihood and integrating  over all $\bm w_n$ and $\bm \lambda_n$, where $\bm \lambda_n$ is the class probability which follows the constructed Dirichlet distribution in EDL.

\vspace{-1mm}
\subsection{Evidential Regression Network}
\label{sec:regression}

Inspired by EDL~\cite{sensoy2018evidential}, which uses a Dirichlet distribution to model the distribution of class probabilities in classification, Evidential Regression Network (ERN)~\cite{amini2020deep} employs a Gaussian distribution to model the predictive outcomes in regression.
Although DER does not originate from subjective logic theory, it shares similar motivations with the vanilla EDL. 
In this part, we elaborate on ERN~\cite{amini2020deep}, its extensions on multivariate~\cite{meinert2021multivariate} and multi-modal~\cite{ma2021trustworthy} data, several improved regularization terms~\cite{ye2024uncertainty,wu2024evidence}, and other related works~\cite{oh2022improving,meinert2023unreasonable,duan2024evidential,pandey2023evidential}.

As the pioneer work, \cite{amini2020deep} explores the basic method of performing DER, which assumes that a target value \(y\) is drawn i.i.d. from the Gaussian distribution $\mathcal{N}(\mu, \sigma^2)$, whose mean \(\mu\) and variance \(\sigma\) follow a Normal Inverse-Gamma (NIG) distribution:
\begin{equation}
	\begin{aligned}
	y &\sim \mathcal{N}(\mu, \sigma^2), \quad
	\mu \sim \mathcal{N}(\gamma, \sigma^2 v^{-1}), \\
	\sigma^2 &\sim \Gamma^{-1}(\alpha, \beta), \quad
	(\mu, \sigma^2) \sim \text{NIG}(\gamma, v, \alpha, \beta),
	\end{aligned}
\end{equation}
where \(\Gamma(\cdot)\) is the gamma function, \(\gamma \in \mathbb{R}\), \(v, \beta \in \mathbb{R}^+\), and \(\alpha > 1\).
Similar to EDL, the parameters of the NIG distribution \(\bm m=(\gamma, v,\alpha, \beta)\) is generated by the neural network, and a Softplus function is employed to ensure the non-negative property of $(v,\alpha,\beta)$ (additional +1 added to \(\alpha\)).
Linear activation is used for the parameter \(\gamma\). 
According to the properties of the NIG distribution, the prediction $\mathbb{E}[\mu]$, aleatoric uncertainty $\mathbb{E}[\sigma^2]$, and epistemic uncertainty $\text{Var}[\mu]$ can be calculated as follows:
\begin{equation}
	\mathbb{E}[\mu] = \gamma, \quad
	\mathbb{E}[\sigma^2] = \frac{\beta}{\alpha - 1}, \quad
	\text{Var}[\mu] = \frac{\beta}{v(\alpha - 1)}.
\end{equation}
Marginalizing over \(\mu\) and \(\sigma^2\), the likelihood of an observation \(y\) given \(\bm m=(\gamma, v,\alpha, \beta)\) can be calculated as:
\begin{equation}
	p(y | \bm m) = \text{St}\left(y; \gamma, \frac{\beta(1 + v)}{v \alpha}, 2 \alpha \right)
\end{equation}
where \(\text{St}(y; \mu_{\text{St}}, \sigma^2_{\text{St}}, \nu_{\text{St}})\) is the Student-t distribution with location \(\mu_{\text{St}}\), scale \(\sigma^2_{\text{St}}\) and degrees of freedom \(\nu_{\text{St}}\).
Using the negative logarithm of the likelihood as the optimization objective and adding a regularization term to minimize incorrect evidence, the loss function of DER is formulated as $\mathcal{L}_{\text{ern}}=\mathcal{L}_{\text{nll}} + \lambda \mathcal{L}_{\text{R}}$, where $\lambda$ is a hyper-parameter.
Specifically, denoting \(\Omega = 2 \beta (1 + v)\), $\mathcal{L}_{\text{nll}}$ and $\mathcal{L}_{\text{R}}$ can be calculated as:
\begin{equation}
	\small
	\begin{aligned}
		\label{L_nll}
		\mathcal{L}_{\text{nll}}
		=& \frac{1}{2} \log \left( \frac{\pi}{v} \right) - \alpha \log(\Omega)
		+ \left( \alpha + \frac{1}{2} \right) \log \left( (y - \gamma)^2 v + \Omega \right) \\
		&+ \log \left( \frac{\Gamma(\alpha)}{\Gamma\left( \alpha + \frac{1}{2} \right)} \right),
		\quad \mathcal{L}_{\text{R}} = |y - \gamma|\cdot(2v+a).
	\end{aligned}
\end{equation}

Furthermore, \cite{meinert2021multivariate} extends the univariate ERN to multivariate regression by replacing the NIG distribution with a normal-inverse-Wishart (NIW) distribution.
Exploring another direction, \cite{ma2021trustworthy} extends the single-modal ERN to multimodal regression by fusing multiple NIG distributions from different modalities into a single NIG distribution. 
Existing works have also explored various additional regularization terms~\cite{ye2024uncertainty,wu2024evidence}.
\cite{ye2024uncertainty} observes that in the high uncertainty area (HUA) of ERN, the gradient of ERN tends to shrink to zero, thereby hindering the correct update of ERN outputs.
To address this issue, this work proposes an uncertainty regularization 
$\mathcal{L}_\text{U} = - | y - \gamma|\cdot\log(\exp(\alpha-1)-1)$.
Similarly, to ensure gradients during evidence contradiction are non-zero, \cite{wu2024evidence} proposes a non-saturating uncertainty regularization:
$\mathcal{L}_\text{U} = (y - \gamma)^2\frac{\nu(\alpha-1)}{\beta(\nu+1)}$.
It is proved that this regularization can ensure that a gradient exists for the prediction throughout the entire domain of definition.
to address the gradient shrinkage problem,  \cite{oh2022improving} proposes incorporating an additional Lipschitz-modified mean squared error loss alongside the existing negative log-likelihood loss.

In addition, existing researches have also explored the ERN method in the following aspects:
\cite{meinert2023unreasonable} provides additional insights into the empirical effectiveness of DER, highlighting its theoretical shortcomings and discussing corrections of how aleatoric and epistemic uncertainties should be extracted.
Aiming to quantify classification uncertainty at the class level, \cite{duan2024evidential} adapts the variance-based approach, widely applied in regression problems, to the classification task. 
Moreover, \cite{pandey2023evidential} extends deep evidential regression~\cite{amini2020deep} to an evidential neural process by incorporating the conditional neural process method~\cite{garnelo2018conditional}.

\vspace{-1mm}
\section{APPLICATIONS}
\label{sec:app}

\subsection{EDL Enhanced Machine Learning}
	\label{sec:machine}
	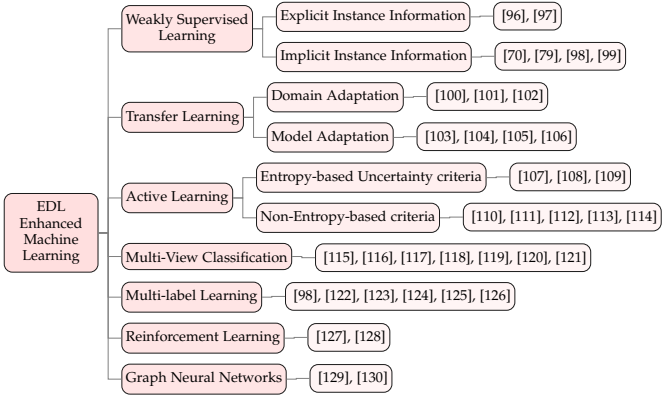
\begin{figure}[t!]
		\centering
		\resizebox{1\linewidth}{!}{
		\begin{forest}
			for tree={
				forked edges,
				grow=east,
				reversed=true,
				anchor=base west,
				parent anchor=east,
				child anchor=west,
				base=middle,
				font=\scriptsize,
				rectangle,
				draw=black, 
				edge=black!50, 
				rounded corners,
				minimum width=2em,
				s sep=5pt,
				inner xsep=3pt,
				inner ysep=2pt
			},
			[EDL Enhanced Machine Learning,leaf4,edge=black!50, fill=mypurple, minimum height=1.2em
			[Weakly Supervised \\ Learning,leaf3,edge=black!50,fill=mypurple, 
			[Explicit Instance Information,leaf2,fill=mypurple,
			[\cite{zhu2022towards,liu2024weakly},leaf1,fill=mypurple]
			]
			[Implicit Instance Information,leaf2,fill=mypurple,
			[\cite{gao2023collecting,gao2023vectorized,chen2022dual,chen2023cascade},leaf1,fill=mypurple]
			]
			]
			[Transfer Learning,leaf3,edge=black!50,fill=mypurple, 
			[Domain Adaptation,leaf2, fill=mypurple,
			[\cite{chen2022evidential,nejjar2024uncertainty,pei2024evidential},leaf1,fill=mypurple]
			]
			[Model Adaptation,leaf2, fill=mypurple,
			[\cite{guo2023bayesian,aguilar2023continual,holmquist2021modeling,linghu2022bayesian},leaf1,fill=mypurple]
			]
			]
			[Active Learning,leaf3,edge=black!50, fill=mypurple, 
			[Entropy-based Uncertainty criteria, leaf2,fill=mypurple,
			[\cite{balaram2022consistency,park2022active,zhang2023revisiting},leaf1,fill=mypurple]
			]
			[Non-Entropy-based criteria, leaf2,fill=mypurple,
			[\cite{hemmer2022deal,sun2022evidential2,shi2020multifaceted,fan2024evidential,chen2024think},leaf1,fill=mypurple]
			]
			]
			[Multi-View Classification,leaf3,edge=black!50, fill=mypurple, 
			[\cite{han2022trusted,xu2022uncertainty,xie2023exploring,zhao2023credible,gao2023reliable,gudhe2022multi,xu2024reliable}, leaf1,fill=mypurple]
			]
			[Multi-label Learning,leaf3,edge=black!50, fill=mypurple, 
			[\cite{ashfaq2023deed,zhao2023towards,gao2023collecting,zhao2023open,zhao2022seed,zhao2023multi}, leaf1,fill=mypurple]
			]
			[Reinforcement Learning,leaf3,edge=black!50, fill=mypurple, 
			[\cite{wang2023deep,yang2023uncertainty}, leaf1,fill=mypurple]
			]
			[Graph Neural Networks,leaf3,edge=black!50, fill=mypurple, 
			[\cite{zhao2020uncertainty,yu2023uncertainty}, leaf1,fill=mypurple]
			]
			]
		\end{forest}
		}
		\caption{An overview of EDL-enhanced machine learning algorithms, including: weakly-supervised learning (section~\ref{WS}), transfer learning  (section~\ref{TL}), active learning (section~\ref{AL}), multi-view classification (section~\ref{MVC}), multi-label learning (section~\ref{ML}), reinforcement learning (section~\ref{RL}), and graph neural networks (section~\ref{GNN}).}
		\label{fig:ML}
		\vspace{-2mm}
	\end{figure}
In the previous sections, we introduced the theoretical development of EDL. Due to its strong usability and extensibility, EDL has also been widely applied to enhance existing classical machine learning paradigms, such as weakly supervised learning and transfer learning. In this section, we introduce the application of EDL in the field of seven typical ML paradigms, as depicted in~\firef{fig:ML}. 

\subsubsection{Weakly Supervised Learning}
\label{WS}
Weakly supervised learning emerges as a crucial paradigm for utilizing extensive datasets that lack precise or comprehensive annotations~\cite{foulds2010review}. Multiple Instance Learning (MIL) is a relatively straightforward and widely applicable paradigm within weak supervision that utilizes Evidential Deep Learning (EDL) for improvement, as both MIL and EDL fall under the category of classification problems. However, EDL was originally proposed for fully-supervised scenarios, where each training instance is individually labeled, providing a fine-grained instance-label pairs. In contrast, the formulation of MIL involves assigning labels $\mathbf{y} \in \mathbb{R}^C$ to sets of instances, known as bags, rather than individual instances~\cite{foulds2010review}. Specifically, for each dimension of the label, $y_i$, it is marked positive if at least one instance within the bag is positive. Since the unavailability of precise instance-level ground truth, EDL is adopted for estimating the ambiguity of instance-level information. 
	
Methods utilizing EDL to measure instance-level ambiguity can be categorized into explicit and implicit approaches based on their optimization differences. Given the collected instance evidence, explicit approaches directly optimize them and derive instance-level predictions, while implicit approaches aggregate instance evidence into bag evidence and predict class probabilities at bag-level instead. 
For explicit approaches, the challenge lies in the mismatch between the predicted results and the given labels.
To select clean positive instance, Zhu et al.~\cite{zhu2022towards} sets thresholds on predicted probabilities and concentration parameters of Dirichlet distribution while Liu et al.~\cite{liu2024weakly} perform weighted summation of the instances in a positive bag to obtain a single pseudo-positive instance.
For implicit approaches, instance-level evidence is predicted first and then aggregated into bag-level evidence, resulting in bag-level categorical distributions. An intuitive scheme for aggregation is to sum the weighted instance-level evidence as follows:
\begin{equation}
	e^{bag}=\sum_{j}^{N}w_{j}e^{instance}_{j}
\end{equation}
Here, $N$ represents the total number of instances in a bag while $e^{bag}$ and $e^{instance}$ denote the bag-level and instance-level evidence, respectively. The instance weights, $w^j$, are either directly predicted by the network~\cite{chen2022dual,gao2023collecting} or determined based on the evidential uncertainty~\cite{gao2023vectorized}. After obtaining the bag-level evidence vector, bag-level categorical distribution and uncertainty can both be derived based on evidential theory. Additionally, \cite{chen2023cascade} proposes a variant, which positions the evidential head after the aggregation operation, i.e., aggregating features instead of evidence.

\vspace{-1mm}
\subsubsection{Transfer Learning}
\label{TL}
Transfer learning aims to help improve the performance of target tasks $\mathcal{T}_T$ in a target domain $\mathcal{D}_T$ using the knowledge in a source domain $\mathcal{D}_S$, acquired by source tasks $\mathcal{T}_S$, where $\mathcal{D}_S\neq\mathcal{D}_T$ or $\mathcal{T}_S\neq\mathcal{T}_T$. This transfer process is typically achieved by reducing the difference between domains, which refers to domain adaptation(DA)~\cite{zhuang2020comprehensive}. 
Traditional DA methods provide simplistic sample prediction, which fails to accurately assess the compatibility between the source domain knowledge and the target domain samples, thereby limiting the fine-grained information transfer. In contrast, evidential networks can capture more comprehensive and specific source domain knowledge in the prediction results of each target sample, enabling more precise domain adaptation~\cite{pei2024evidential, chen2022evidential, zhang2023revisiting}. 
Along with domain adaptation, some studies focus on transferring distributional knowledge from a source model to a target model within the same domain, such as knowledge distillation~\cite{guo2023bayesian,aguilar2023continual,holmquist2021modeling} and pre-trained model fine-tuning~\cite{linghu2022bayesian}, which we refer to as model adaptation. For clarity, we will also discuss how evidential theory guides model adaptation in this section.

For EDL-enhanced DA, Chen et al.~\cite{chen2022evidential} explore the Universal Domain adaptation (UniDA) setting which expands upon unsupervised DA by introducing the concept of category shift. 
UniDA poses the technical challenge of estimating the label distribution on the target domain and detecting potential target “unknown” (unique) samples. To this end, \cite{chen2022evidential} proposes using EDL to construct category-aware thresholds $\delta_j$ with total evidence $S^t$ and concentration parameters $\boldsymbol{\alpha}^t$ to reject target "unknown" samples, which is defined as:
\begin{equation}
	\small
	y^t=\begin{cases}
		j & \text{if } logS^t\geq\delta_j, j=\arg\max_{1\leq k\leq L_s}\alpha_k^t \\
		unknown & \text{if } logS^t < \delta_j, j=\arg\max_{1\leq k\leq L_s}\alpha_k^t
	\end{cases}
\end{equation}
Here, $y^t$ denotes the predicted label of a target sample while $L_s$ denotes the number of labels within source domain. 
Another work is for Unsupervised Domain Adaptation in Regression (UDAR)~\cite{cortes2011domain}, which involves transforming the classification tasks in UDA into regression tasks. Unlike the clustered structure in classification tasks, the embedded space structure in regression tasks is typically not apparent. Therefore, Nejjar et al.~\cite{nejjar2024uncertainty} propose aligning the parameters of higher-order evidential distributions, which provide a comprehensive reflection of the entire embedding space:
\begin{equation}
	\mathcal{L}_{align}=\sum_{k=1}^{N_s}\phi(\mathbf{m^s_k})-\sum_{j=1}^{N_t}\phi(\mathbf{m^t_j})
\end{equation}
Here, $\phi$ is the criterion to measure the discrepancy between source and target evidence vectors, such as Maximum Mean Discrepancy \cite{borgwardt2006integrating}. $\mathbf{m^s}$ and $\mathbf{m^t}$ represents the parameter vectors forming evidential distributions. 
For Multi-Source-Free Unsupervised Domain Adaptation (MSFDA)~\cite{ahmed2021unsupervised} which aims to transfer multiple source models to a target domain without access to source data, Pei et al.~\cite{pei2024evidential} propose a instance-customized source model aggregation strategy through sample preference. Formally, the sample preference is defined as $Sp=max_i e_i+1$, which describes the preferences of target samples for different source models and enable fine-grained source model aggregation.

The following works, focused on model adaptation, involve two tasks: Knowledge Distillation (KD) and Few-Shot Classification (FSC). 
In the process of knowledge distillation, the student network is typically an evidential network, while the teacher network can be either a complex Bayesian neural network~\cite{guo2023bayesian} or an evidential network~\cite{aguilar2023continual, holmquist2021modeling}. Due to the introduction of uncertainty by the evidential network, it is possible to align not only the predicted posterior distribution but also the predicted uncertainty: 
\begin{equation}
	\mathcal{L}_{KD}=\mathcal{D}(P_s||P_t) + \lambda\mathcal{D}(U_s||U_t)
\end{equation}
Here, $\mathcal{D}$ represents the metric used to measure the difference between the teacher and student predictions, such as KL divergence.
For Few-Shot Classification (FSC), Linghu et al.~\cite{linghu2022bayesian} utilize the unique mathematical properties of the Dirichlet distribution to achieve model fine-tuning by combining the meta-trained model parameters with those of the pre-trained model.

\vspace{-1mm}
\subsubsection{Active Learning}
\label{AL}
Active Learning (AL) aims to achieve higher accuracy with fewer labeled training instances by strategically selecting the data from which it learns~\cite{ren2021survey}. 
In this learning process, uncertainty is often used as a criterion for selecting samples that need to be labeled. 
EDL is distinguished by its efficiency in performing uncertainty reasoning with a single forward propagation. More importantly, the use of Subjective Logic (SL)~\cite{josang2016subjective} allows for a more nuanced categorization of uncertainty, leading to diverse uncertainty-based querying strategies. In addition to computing epistemic uncertainty and aleatoric uncertainty, 
we can further decompose uncertainty into vacuity and dissonance, providing greater flexibility in sample selection. 

Given the predicted class distribution and its conjugate prior, epistemic and aleatoric uncertainty defined by entropy are mostly used as the criteria for sample selection. 
Balaram et al.~\cite{balaram2022consistency} estimate the aleatoric uncertainty (AU) which arises from inherent noise in the data, through the expected entropy of the data distribution $P(y|p)$: 
\begin{equation}
	AU=\mathbb{E}_{p\sim Dir(\mathbf{\alpha})}\{\mathcal{H}[P(y|p)]\}
	\label{AU}
\end{equation}
Here, $\mathcal{H}$ denotes the Shannon entropy, and $p$ follows a Dirichlet distribution.
Park et al. \cite{park2022active} compute the epistemic uncertainty (EU)
as the mutual information between the label $y$ and its categorical distribution $p$:
\begin{equation}
	EU=\mathcal{H}[\mathbb{E}_{p\sim Dir(\mathbf{\alpha})}[P(y|p)]]-\mathbb{E}_{p\sim Dir(\mathbf{\alpha})}[\mathcal{H}[P(y|p)]]
	\label{EU}
\end{equation}
Since both AU and EU reflect the informativeness of samples, Zhang et al.~\cite{zhang2023revisiting} combine them with different weights. 

Apart from the entropy-based sample selection criteria, Hemmer et al.~\cite{hemmer2022deal} use the difference between the highest two predicted probabilities to construct the margin for sampling unlabeled samples. Additionally, while the epistemic uncertainty in Equation~\ref{edl-probability} is a feasible criterion~\cite{sun2022evidential2, fan2024evidential}, it only reflects the vacuity of evidence. According to SL~\cite{josang2016subjective}, another dimension of epistemic uncertainty, termed dissonance, reflects the consistency of the collected evidence. 
Considering that vacuity and dissonance reflect the extent of a model's lack of knowledge and the degree of evidence conflict, respectively, Shi et al.~\cite{shi2020multifaceted} combine these two metrics using a time-dependent coefficient. 
Different from the above methods, Chen et al.~\cite{chen2024think} utilize epistemic uncertainty to calibrate the aleatoric uncertainty for AL.


\vspace{-1mm}
\subsubsection{Multi-View Classification}
\label{MVC}
Multi-view classification has emerged as a prominent research topic involves leveraging information from multiple data sources or "views" to enhance the classification models. However, progress in this field is hampered by the varying quality of views~\cite{li2018survey}. Recent advancements in EDL have shed lights on the multi-view classification, offering robust fusion strategies and decision explainability. 

Trusted Multi-view Classification (TMC), a pioneering framework proposed by Han et al.~\cite{han2022trusted}, encompasses three pivotal steps for dynamically integrating different views: reliable single-view Dirichlet distributions are obtained via variational approximation; these Dirichlet distributions induce subjective opinions; Dempster-Shafer Theory (DST)-based combination rule is adopted to integrate these subjective opinions. Given two subjective opinions $w^1=\{\{b_i^1\}_{i=1}^C,u^1\}$ and $w^2=\{\{b_i^2\}_{i=1}^C,u^2\}$, the combination $w=\{\{b_i\}_{i=1}^C,u\}$ is computed as follows:
\begin{equation}
	\small
	b_c=\frac{1}{1-Conf}(b_i^1b^2_i+b^1_iu^2+b_i^2u^1),\ u=\frac{1}{1-Conf}u^1u^2
	\label{DSrule}
\end{equation}
TMC successfully introduces evidential deep learning into multi-view classification tasks, paving the way for subsequent research in multi-view related work~\cite{gao2023reliable,xie2023exploring,zhao2023credible,gudhe2022multi, xu2024reliable}. Xie et al. \cite{xie2023exploring} impute the incomplete multi-view data and integrate the imputed data with evidential theory.
Despite the significant success of DST theory, its applicability is limited as it assumes consistency in multi-view data and overlooks complementarity. Therefore, Xu et al.~\cite{xu2022uncertainty} explicitly model consistent and complementary relations through a degradation layer to learn the mappings from the fused evidence to view-specific evidences. 

\vspace{-1mm}
\subsubsection{Multi-label Learning}
\label{ML}
In the real world, images/objects often possess multiple semantic attributes simultaneously, presenting the paradigm of multi-label learning. 
While Evidential Deep Learning was originally developed for the single-label classification~\cite{sensoy2018evidential}, it can be easily extended to multi-label scenarios by leveraging the decomposition concept of Binary Relevance~\cite{zhang2013review}.

According to Binary Relevance, a $C$-way multi-label classification problem is decomposed into $C$ independent binary classification problems. The multinomial distribution and evidential priors over likelihood function are replaced by Bernoulli distribution and Beta distributions, respectively. Accordingly, the multi-label optimization objective $\mathcal{L}_{ML}$ turns out to be the following form:
\begin{equation}
	\mathcal{L}_{ML}=\int[\sum_{i=1}^C L(y_i|p_i) Beta(p_i|\alpha_i,\beta_i)]
\end{equation}
Here, $L(y_i|p_i)$ represents the loss function, such as cross-entropy loss~\cite{gao2023collecting} and type-II maximum likelihood~\cite{ashfaq2023deed}. $\alpha_i$ and $\beta_i$ represent the parameters that constitute the Beta function for class $i$.
It is worth noting that, given the input embeddings, the evidential heads for each category are different, thus the categories are conditionally independent.

Although the above binary decomposition framework has been adopted by many studies due to its simplicity and ease of understanding~\cite{gao2023collecting,zhao2023open,zhao2022seed,zhao2023multi,ashfaq2023deed,zhao2023towards}, its drawbacks are also evident. Firstly, similar to Binary Relevance, this method may suffer from data imbalance issue. 
Secondly, this method lacks a solution for cases where the predicted set of classes is empty. 

\vspace{0mm}
\subsubsection{Reinforcement Learning}
\label{RL}
Deep reinforcement learning differs from other ML algorithms because it employs neural networks to tackle sequential decision-making problems requiring agent-environment interactions. At each time step $t$, an agent observes a state $s_t$, then interacts with the environment by taking an action $a_t$ based on its policy $\pi$. Subsequently, the environment transitions to another state $s_{t+1}$ and provides a reward $r_{t+1}$ to the agent. The goal of the agent is to learn the optimal policy that maximizes the expected cumulative reward~\cite{arulkumaran2017deep}. To achieve this goal, the agent needs to balance exploitation and exploration of the environment, which can be accomplished by incorporating evidential uncertainty into the reward. Conversely, uncertainty-based reward can serve as guidance for prediction risk and model confidence calibration without the need of ground truth. 

Evidential policy networks is crucial for quantifying uncertainty and forms the basis for designing uncertainty-based reward functions. A general uncertainty-aware reward function $r^e_{t+1}$ when taking an action $a_t$ is: 
\begin{equation}
	r^e_{t+1}(s_t,a_t,s_{t+1})=r_{t+1}(s_t,a_t,s_{t+1}) + U(\pi^e)
\end{equation}
where $U(\pi^e)$ denotes the uncertainty reward quantified by evidential policy network $\pi^e$. Wang et al.~\cite{wang2023deep} instantiate $U(\pi^e)$ as epistemic uncertainty to encourage the agent to explore unknown areas. Given pre-trained EDL backbones, 
Yang et al.~\cite{yang2023uncertainty} design rewards which quantifies how closely the predicted uncertainty is aligned with the model prediction risk. 

\vspace{0mm}
\subsubsection{Graph Neural Networks}
\label{GNN}
{
For graph structure data,~\cite{zhao2020uncertainty} designs a Graph-based Kernel Dirichlet distribution Estimation (GKDE) method for 
OOD nodes detection.
First, a subjective graph neural network (S-GNN) $f$ is designed to construct the node-level Dirichlet distribution $\text{Dir}(\bm{p}_i|\bm{\alpha}_i)$ for the node $i$, where $\bm{\alpha}_i = f_i(A, \bm{r}; \bm{\theta})$.
$f_i$ is the output for node $i$, $\bm r$ is the node-level feature matrix, $\bm{\theta}$ is the model parameters, and $A$ is the adjacency matrix.
Note that S-GNN differs from traditional GNNs by replacing the softmax layer, which typically outputs class probabilities, with an evidence function. 
Subsequently, 
each training node is treated as evidence for its corresponding class label. To improve the OOD detection performance, Yu et al.~\cite{yu2023uncertainty} propose two uncertainty-aware regularization terms for evidential GNNs.

\vspace{0mm}
\subsubsection{Discussion}
EDL exhibits the following advantages: (1) Multi-scenario adaptation. The previous subsections have discussed the applications of EDL in various machine learning paradigms. From labeled data to unlabeled data, from single-domain to multi-domain, from single-round learning to multi-round learning, from single-view classification to multi-view integration, from single-label learning to multi-label learning, and from static learning to interactive learning, EDL demonstrates robust adaptability. (2) Multi-source uncertainty quantification. Benefiting from the Dirichlet distribution, multi-source uncertainty has a closed-form expression, including epistemic and aleatoric uncertainties~\cite{park2022active,sun2022evidential2, cai2023evora,balaram2022consistency}. Additionally, based on evidential theory, more specific and contextual uncertainties have also become possible, such as decomposing uncertainty into vacuity and dissonance~\cite{shi2020multifaceted}, and sample preference for a specific source domain~\cite{pei2024evidential}. (3) Fine-grained information mining. The introduction of evidential theory provides a credibility perspective, which motivates the exploration of more fine-grained information, such as the credibility of instances~\cite{zhu2022towards,gao2023collecting,gao2023vectorized,chen2022dual}, alignment of distributions between domains~\cite{cao2019learning,nejjar2024uncertainty,guo2023bayesian,holmquist2021modeling}, measurement of sample informativeness~\cite{park2022active,shi2020multifaceted}, and more balanced exploration of the environment \cite{wang2023deep,guo2017calibration}.

Additionally, evidential machine learning for ML is expected to be further studied in the following two aspects: (1)  {Employment of advanced methodologies}. In Section~\ref{sec:the_exp}, various theoretical improvements to vanilla EDL are flourishing~\cite{chen2023r,deng2023uncertainty,oh2022improving,hu2021multidimensional}, but few works adapt these improved frameworks to machine learning algorithms. Research on EDL is currently in its early stages, with theoretical and practical aspects advancing concurrently, requiring further exploration. A more general and effective evidential framework is still and eagerly anticipated. (2) {Applications for more ML settings}.  EDL has been less explored in certain machine learning paradigms such as unsupervised learning and federated learning. 

\vspace{-1mm}
\subsection{EDL in Downstream Applications}
\label{sec:downstream}
The application of EDL in downstream tasks is flourishing, encompassing a wide range of tasks across multiple domains. As shown in~\firef{fig:D_APP}, our survey covers six domains: computer vision, natural language processing, cross-modal learning, autonomous driving, tasks related to open-world scenarios, and scientific fields. A detailed summary of the applications of EDL can be found in Table~\ref{tab_app}.
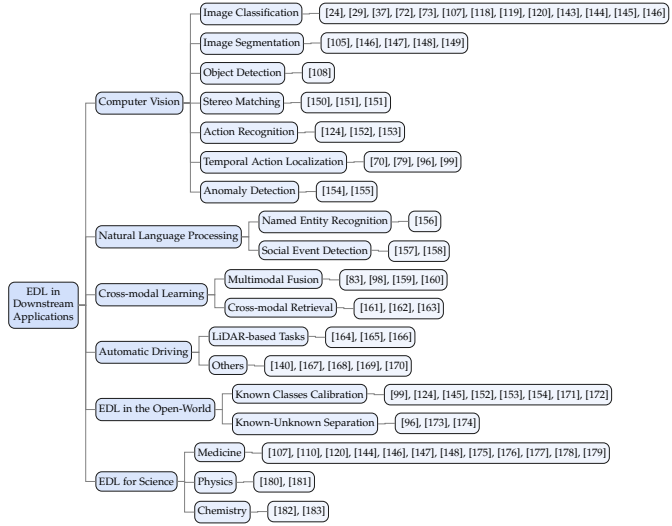
\begin{figure}[t!]
\centering
\resizebox{1\linewidth}{!}{
	\begin{forest}
		for tree={
			forked edges,
			grow=east,
			reversed=true,
			anchor=base west,
			parent anchor=east,
			child anchor=west,
			base=middle,
			font=\scriptsize,
			rectangle,
			draw=black, 
			edge=black!50, 
			rounded corners,
			minimum width=2em,
			s sep=5pt,
			inner xsep=3pt,
			inner ysep=2pt
		},
		[EDL in Downstream Applications,leaf4,edge=black!50, fill=mygreen, minimum height=1.2em
		[Computer Vision,leaf3,edge=black!50, fill=mygreen, minimum height=1.2em,
		[Image Classification ,leaf2,fill=mygreen,
		[\cite{sensoy2020uncertainty,zhao2019quantifying,hu2021multidimensional,deng2023uncertainty,chen2023r,tong2021fusion,gao2023reliable,gudhe2022multi,balaram2022consistency,ghesu2019quantifying,ji2024spectral,zhao2023credible,ren2023uncertainty},leaf1,fill=mygreen]
		]
		[Image Segmentation,leaf2,fill=mygreen,
		[\cite{holmquist2021modeling,zou2022tbrats,li2023region,ren2023uncertainty,shi2024evidential},leaf1,fill=mygreen]
		]
		[Object Detection,leaf2,fill=mygreen,
		[\cite{park2022active},leaf1,fill=mygreen]
		]
		[Stereo Matching,leaf2,fill=mygreen,
		[\cite{wang2022uncertainty,lou2023elfnet,lou2023elfnet},leaf1,fill=mygreen]
		]
		[Action Recognition,leaf2,fill=mygreen,
		[\cite{bao2021evidential,zhang2023learning,zhao2023open},leaf1,fill=mygreen]
		]
		[Temporal Action Localization,leaf2,fill=mygreen,
		[\cite{zhu2022towards,chen2023cascade,chen2022dual,gao2023vectorized},leaf1,fill=mygreen]
		]
		[Anomaly Detection,leaf2,fill=mygreen,
		[\cite{bao2022opental,sun2022evidential1},leaf1,fill=mygreen]
		]
		]
		[Natural Language Processing,leaf3,edge=black!50, fill=mygreen, minimum height=1.2em,
		[Named Entity Recognition,leaf2,fill=mygreen,
		[\cite{zhang2023ner},leaf1,fill=mygreen]
		]
		[Social Event Detection,leaf2,fill=mygreen,
		[\cite{ren2022evidential,ren2023uncertaintyevent},leaf1,fill=mygreen]
		]
		]
		[Cross-modal Learning,leaf3,edge=black!50, fill=mygreen, minimum height=1.2em,
		[Multimodal Fusion,leaf2,fill=mygreen,
		[\cite{gao2023collecting,shao2024dual,ma2021trustworthy,xu2022deep},leaf1,fill=mygreen]
		]
		[Cross-modal Retrieval,leaf2,fill=mygreen,
		[\cite{qin2022deep,li2024prototype,li2023dcel},leaf1,fill=mygreen]
		]
		]
		[Automatic Driving,leaf3,edge=black!50, fill=mygreen, minimum height=1.2em,
		[LiDAR-based Tasks,leaf2,fill=mygreen,
		[\cite{bauer2019deep,liu2021efficient,ali2023delo},leaf1,fill=mygreen]
		]
		[Others, leaf2,fill=mygreen,
		[\cite{petek2022robust,li2023distil,cai2023evora,itkina2023interpretable,zhang2023trep},leaf1,fill=mygreen]
		]
		]
		[EDL in the Open-World,leaf3,edge=black!50, fill=mygreen, minimum height=1.2em,
		[Known Classes Calibration,leaf2,fill=mygreen,
		[\cite{bao2021evidential,chen2023cascade,sapkota2023adaptive,bao2022opental,ji2024spectral,zhang2023learning,zhao2023open,wang2024towards},leaf1,fill=mygreen]
		]
		[Known-Unknown Separation, leaf2,fill=mygreen,
		[\cite{su2023hsic,zhu2022towards,yu2023adaptive},leaf1,fill=mygreen]
		]
		]
		[EDL for Science,leaf3,edge=black!50, fill=mygreen, minimum height=1.2em,
		[Medicine,leaf2,fill=mygreen,
		[\cite{gudhe2022multi,ghesu2019quantifying,hemmer2022deal, balaram2022consistency,xu2023eviprompt,zou2022tbrats,li2023region,ren2023uncertainty,zheng2023evidential,zhu2022personalized, zhu2022enhancing, zhu2022iomt},leaf1,fill=mygreen]
		]
		[Physics ,leaf2,fill=mygreen,
		[\cite{koh2021evaluating,tan2023single},leaf1,fill=mygreen]
		]
		[Chemistry,leaf2,fill=mygreen,
		[\cite{soleimany2021evidential,vazquez2024outlier}, leaf1, fill=mygreen]
		]            
		]
		]
	\end{forest}
}
\caption{An overview of downstream applications of EDL, including six fields: computer vision (section~\ref{CV}), natural language processing (section~\ref{NLP}), cross-modal learning (section~\ref{CM}), automatic driving (section~\ref{AD}), EDL in the open-world (section~\ref{OS}) and EDL for science (section~\ref{SCI})}
\vspace{-2mm}
\label{fig:D_APP}
\end{figure}

\vspace{-1mm}
\subsubsection{Computer Vision}
\label{CV}
EDL has significantly enhanced the interpretability and reliability of models in analyzing visual content, proving instrumental in navigating the complexities and ambiguities inherent in both static images and dynamic video streams. 

Image classification is the most representative and fundamental task to validate the effectiveness and generalizability of EDL. From an effectiveness standpoint, image classification tasks, particularly Out-Of-Distribution (OOD) detection, have been pivotal for validating evidential networks since its inception~\cite{sensoy2020uncertainty}. Subsequent theoretical advancements, demonstrated in Section~\ref{sec:the_exp}, have consistently utilized this task as a fundamental benchmark for evaluating model performance~\cite{sensoy2020uncertainty,zhao2019quantifying,hu2021multidimensional,deng2023uncertainty,chen2023r,tong2021fusion,yue2022three}. In addition to natural images used for theoretical exploration, recent works expand the scope to specialized images. In the medical field, evidential theory has been leveraged for automated diagnosis, including liver fibrosis staging~\cite{gao2023reliable}, screening mammograms assessment~\cite{gudhe2022multi}, and radiograph classification~\cite{balaram2022consistency,ghesu2019quantifying}. Similarly, in remote sensing, the classification of hyperspectral images~\cite{ji2024spectral,yu2023uncertainty} and aerial-ground dual-view images~\cite{zhao2023credible} are included.

In addition to image classification, object detection and segmentation are two of the most fundamental tasks in computer vision. For detection, although the objective entails both object classification and localization, current works~\cite{su2023hsic,park2022active} simply apply EDL as a classifier. For instance, Su et al.~\cite{su2023hsic} classify the proposals generated by the backbone network (Faster R-CNN) with vanilla evidential network. Park et al. \cite{park2022active} replace the ReLU evidential function with a a Softmax function for achieving a sharp resultant distribution and confident prediction. For image segmentation, akin to image classification in categorizing pixels within an image, we can seamlessly extends the application of evidential theory. Nevertheless, the shift in prediction objectives somewhat compromises the model's classification performance. Therefore, current research aims to improve the performance of evidential models in segmentation tasks by learning pixel-aware uncertainties~\cite{ren2023uncertainty, shi2024evidential} and designing uncertainty-aware objectives~\cite{zou2022tbrats,li2023region,sirohi2023uncertainty}.

Stereo matching strives to estimate the disparity between the given stereo pair of each pixel. Although stereoscopic matching is a combination of classification and regression problems, existing approaches~\cite{lou2023elfnet,wang2022uncertainty} predominantly focuses on the uncertainties in the regression aspect. Specifically, a normal distribution with mean and variance following a Normal Inverse-Gamma (NIG) distribution is placed on pixel-wise disparity, followed with various strategies such as using the multi-scale cost volume information~\cite{lou2023elfnet} or gradient variation of evidence parameters~\cite{wang2022uncertainty}.


For video-related tasks, current research on EDL primarily focuses on the recognition and localization/detection areas. 
The commonly used action recognition pipeline involves feature extractor from input data using a backbone network, followed by a evidential classifier~\cite{bao2021evidential,zhang2023learning,zhao2023open}. To fully leverage the advantages of evidential theory, current works are not confined to the simplest closed-set, single-label action recognition task settings. Instead, they focus on more challenging open-set~\cite{bao2021evidential,zhang2023learning,zhao2023open} and multi-label~\cite{zhao2023open} action recognition scenarios, which is discusses in Section~\ref{OS} and Section~\ref{ML}. Temporal action localization demands accurate action categorization and precise temporal localization within videos. Methods categorize into those treating these goals independently~\cite{bao2022opental} and those integrating both into unified solutions~\cite{chen2023cascade,chen2022dual,gao2023vectorized}. Video anomaly detection (VAD) can be defined as a technique meant for identifying the abnormal patterns or trends present in the data. Besides the intuitive binary classification formulation~\cite{zhu2022towards}, Sun et al.~\cite{sun2022evidential1} encode multiple visual cues and minimize the energy of the Gaussian Mixture Model (GMM) to obtain more effective evidence.

\subsubsection{Natural Language Processing}
\label{NLP}
Quantifying the uncertainty of neural networks is becoming a critical research direction for natural language processing. Though EDL was initially proposed based on image data, its fundamental concept for uncertainty quantification is independent of the modality. Consequently, tasks like named entity recognition (NER)~\cite{zhang2023ner}, which involves classifying each word in an input word sequence, and social event detection (SED)~\cite{ren2022evidential,ren2023uncertaintyevent}, which categorizes social messages, are well-suited for incorporating evidential uncertainty. The vanilla EDL method fails to achieve satisfactory results for both NER and SED due to a common issue: imbalanced data distribution. To address this issue, Zhang et al.~\cite{zhang2023ner} replace the one-hot ground truth labels with importance weights leading to a re-distributed model attention.
Additionally, Ren et al.~\cite{ren2023uncertaintyevent} approach the problem from the perspective of optimizing the latent space by incorporating a margin related to class uncertainty in the regularization term to separate ambiguous boundaries.


\subsubsection{Cross-modal Learning}
\label{CM}
Currently, research applying evidential deep learning (EDL) to cross-modal data is focused on two primary topics: multimodal fusion and cross-modal retrieval. 
Multimodal fusion aims to integrate multiple modalities into a cohesive whole at either the feature level~\cite{gao2023collecting} or the decision level~\cite{shao2024dual,ma2021trustworthy,xu2022deep}, leveraging their complementary strengths.
In contrast, cross-modal retrieval aims to model the correspondence between modalities, with EDL being used to estimate the uncertainty in cross-modal alignments~\cite{qin2022deep,li2024prototype,li2023dcel}.


For the feature-level, we can simply use additive operation~\cite{gao2023collecting} for evidential multimodal fusion. Decision-level fusion refers to the process of integrating the subjective opinions belonging to different modalities~\cite{xu2024reliable}. Notably, if each modality is considered a view, this strategy shares the same theoretical foundation with Section~\ref{MVC}~\cite{shao2024dual,xu2022deep,chang2022fast}. Moreover, Liu et al.~\cite{liu2023integrating} apply the fusion rule in~\cite{ma2021trustworthy} to the multimodal named entity recognition task, where they fused the NIG distributions obtained from the text and image modalities to arrive at a unified decision.


In cross-modal retrieval, the objective is to rank the relevance between a query and each of $N$ elements in the complementary set. Obviously, the task can be viewed as an $N$-way classification problem, where the complementary set corresponds to the typical category set. It should be noted that cross-modal retrieval is fundamentally a bidirectional classification problem which leads to bidirectional evidence.
Given the bidirectional evidence, Qin et al.~\cite{qin2022deep} isolate mismatched pairs with noisy correspondence while Li et al.~\cite{li2024prototype} calibrate the predictions with data uncertainty. 

\begin{table*}[!t]
	\small
		\centering
		\caption{\footnotesize A detailed summary of the applications of EDL. Note that $L_{mse-edl}$, $L_{ce-edl}$, $L_{nll-edl}$, and $L_{nll}$ correspond to Equation~\ref{edl-loss}, \ref{edl-loss-ce}, \ref{edl-loss-nll}, \ref{L_nll} and KL Regularization represents Equation~\ref{kl-divergence}. Additional Loss refers to those outside the vanilla EDL framework, while Uncertainty Estimation pertains to the methods of uncertainty quantification used in the literature.}
		\label{tab_app}
		\vspace{-2mm}
		\resizebox{\textwidth}{!}{
			\begin{tabular}{cccccccc}
				\toprule
				Publication venue & ML Paradigm & Downstream Task & Loss Type & Evidential Function & KL Regularization & Uncertainty Estimation & Additional Loss \\ \midrule
				NeurIPS2020\cite{shi2020multifaceted} & Active Learning & Image Classification & $L_{mse-edl}$ & ReLU & \ding{55} & $u=\frac{C}{S}$,  $\sum_{i\in\mathbb{X}}\left(\frac{\bm b_i\sum_{j\in\mathbb X\backslash i}\bm b_j\text{Bal}(j,i)}{\sum_{j\in\mathbb X\backslash i}\bm b_j}\right)$ & \checkmark \\ \midrule
				CVPR2021~\cite{bao2021evidential} & Supervised Learning & Open Set Action Recognition & $L_{nll-edl}$ & Exp & \ding{55} & $u=\frac{C}{S}$ & \checkmark \\ \midrule
				ICRA2021~\cite{liu2021efficient} & Supervised Learning & LiDAR-based Navigation & $L_{ern}$ & Softplus & \ding{55} & $Var[\mu]=\frac{\beta}{v(\alpha-1)}$ & \checkmark \\ \midrule
				NeurIPS-W2021~\cite{koh2021evaluating} & Supervised Learning & Particle Classification & $L_{nll-edl}$ \& $L_{ce-edl}$ \&  $L_{mse-edl}$ & ReLU & \ding{55} & —— & \ding{55} \\ \midrule
				ACS Cent. Sci.2021~\cite{soleimany2021evidential} & Supervised Learning & Molecular Structure-property Prediction & $L_{ern}$ & Softplus & \ding{55} & $Var[\mu]=\frac{\beta}{v(\alpha-1)}$ & \checkmark \\ \midrule
				ECCV2022~\cite{chen2022dual} & Weakly-supervised Learning & Temporal Action Localization & $L_{nll-edl}$ & ReLU & $\times$ & $u=\frac{C}{S}$ & \checkmark \\ \midrule
				ECCV2022~\cite{zhu2022towards} & Weakly-supervised Learning & Open Set Video Anomaly Detection & $L_{nll-edl}$ & ReLU & \ding{55} & $u=\frac{C}{S}$ & \checkmark \\ \midrule
				AAAI2022~\cite{chen2022evidential} & Transfer Learning & Universal Domain Adaptation & $L_{nll-edl}$ & Exp & \ding{55} & $u=\frac{C}{S}$ & \checkmark \\ \midrule

				TPAMI2022~\cite{han2022trusted} & Multi-View Classification & Image Classification & $L_{ce-edl}$ & ReLU & \checkmark & $u=\frac{1}{1-C}u^1u^2$ & \ding{55} \\ \midrule
				MM2022~\cite{sun2022evidential1} & Supervised Learning & Video Anomaly Detection & —— & ReLU & \ding{55} & $u-max_ip_i$ & \ding{55} \\ \midrule
				CVPR2022~\cite{bao2022opental} & Supervised Learning & Open Set Temporal Action Localization & $L_{nll-edl}$ & Exp & \ding{55} & $u=\frac{C}{S}$ & \checkmark \\ \midrule
				ICWS2022~\cite{ren2022evidential} & Supervised Learning & Social Event Detection & $L_{ce-edl}$ & ReLU & \checkmark & $u=\frac{C}{S}$ & \checkmark \\ \midrule
				MM2022~\cite{qin2022deep} & Supervised Learning & Cross-modal Retrieval & $L_{mse-edl}$ & Exp \& Tanh & \checkmark & $u=\frac{C}{S}$ & \checkmark \\ \midrule
				CVPR2023~\cite{chen2023cascade} & Weakly-supervised Learning & Open-world Temporal Action Localization & $L_{nll-edl}$ & Exp & \ding{55} & $u=\frac{C}{S}$ & \checkmark \\ \midrule
				CVPR2023~\cite{gao2023collecting} & Weakly-supervised Learning & Audio-Visual Event Perception & $L_{ce-edl}$ & Exp & \ding{55} & $u=\frac{C}{S}$ & \checkmark \\ \midrule
				TPAMI2023~\cite{gao2023vectorized} & Weakly-supervised Learning & Temporal Action Localization & $L_{nll-edl}$ & ReLU & \ding{55} & $u=\frac{C}{S}$ & \checkmark \\ \midrule
				ICCV2023~\cite{aguilar2023continual} & Transfer Learning & Incremental Object Classification & $L_{nll-edl}$ & Softmax & \checkmark & $u=\frac{C}{S}$,  $\sum_{i\in\mathbb{X}}\left(\frac{\bm b_i\sum_{j\in\mathbb X\backslash i}\bm b_j\text{Bal}(j,i)}{\sum_{j\in\mathbb X\backslash i}\bm b_j}\right)$ & \checkmark \\ \midrule
				ICLR2023~\cite{park2022active} & Active Learning & Object Detection & $L_{ce-edl}$ & Softmax & \ding{55} & $I[y,p]$ & \checkmark \\ \midrule
				ICLR2023~\cite{sun2022evidential2} & Active Learning & Scene Graph Generation & $L_{mse-edl}$ & ReLU & \ding{55} & $u=\frac{C}{S}$ & \checkmark \\ \midrule
				CVPR2023~\cite{xie2023exploring} & Multi-View Classification & Incomplete Multi-View Classification & $L_{ce-edl}$ & Softplus & \checkmark & $u=\frac{1}{1-C}u^1u^2$ & \ding{55} \\ \midrule
				AI2023~\cite{ashfaq2023deed} & Multi-label Learning & Diagnosis Prediction & $L_{ce-edl}$ & ReLU & \ding{55} & $Var[Dir(p)]$ & \checkmark \\ \midrule
				CVPR2023~\cite{zhao2023open} & Multi-label Learning & Open Set Action Recognition & $L_{ce-edl}$ & ReLU & \ding{55} & $u=\frac{C}{S}$ & \ding{55} \\ \midrule
				NeurIPS2023~\cite{yang2023uncertainty} & Reinforcement Learning & Scene Segmentation & $L_{ce-edl}$ & Softplus & \ding{55} & $\frac{\alpha_i}{\sum_k\alpha_k}, \sum_k\alpha_k$ & \ding{55} \\ \midrule
				ICML2023~\cite{wang2023deep} & Reinforcement Learning & Unique Behavioral Pattern Discovery & $L_{ern}$ & Softplus & \ding{55} & $Var[\mu]=\frac{\beta}{v(\alpha-1)}$ & \checkmark \\ \midrule
				CVPR2023~\cite{lou2023elfnet} & Supervised Learning &  Stereo Matching & $L_{ern}$ & Softplus & \ding{55} & $Var[\mu]=\frac{\beta}{v(\alpha-1)}$ & \checkmark \\ \midrule
				MM2023~\cite{su2023hsic} & Few-Shot & Open-Set Object Detection & $L_{ce-edl}$ & Exp & \ding{55} & $u=C/S$ & \checkmark \\ \midrule
				MM2023~\cite{zhang2023learning} & Supervised Learning & Open Set Action Recognition & $L_{nll-edl}$ & ReLU & \ding{55} & $u=\frac{C}{S}$ & \checkmark \\ \midrule
				TKDE2023~\cite{ren2023uncertaintyevent} & Supervised Learning & Social Event Detection & $L_{ce-edl}$ & ReLU & \checkmark & $u=\frac{C}{S}$ & \checkmark \\ \midrule
				arXiv2023~\cite{zhang2023ner} & Supervised Learning & Named Entity Recognition & $L_{ce-edl}$ & Softplus & \checkmark & $u=\frac{C}{S}$ & \checkmark \\ \midrule
				MM2023~\cite{li2023dcel} & Supervised Learning & Text-Based Person Retrieval & $L_{nll-edl}$ & ReLU \& Tanh & \checkmark & $u=\frac{C}{S}$ & \checkmark \\ \midrule
				NeurIPS2023~\cite{li2024prototype} & Supervised Learning & Cross-modal Retrieval & —— & Similarity & \ding{55} & $u=1-\frac{C}{S}$ & \checkmark \\ \midrule
				ICLR2023~\cite{sapkota2023adaptive} & Supervised Learning & Open Set Detection & $L_{mse-edl}$ & ReLU & \checkmark & $u=\frac{C}{S}$ & \ding{55} \\ \midrule
				ICML2023~\cite{zheng2023evidential} & Active Learning & Medical Image Captioning & $L_{nll-edl}$ & ReLU & \checkmark & $u=\frac{C}{S}$,  $\sum_{i\in\mathbb{X}}\left(\frac{\bm b_i\sum_{j\in\mathbb X\backslash i}\bm b_j\text{Bal}(j,i)}{\sum_{j\in\mathbb X\backslash i}\bm b_j}\right)$ & \ding{55} \\ \midrule
				CVPR2024~\cite{zhang2023revisiting} & Transfer Learning/Active Learning & Multi-source Active Domain Transfer & $L_{nll-edl}$ & Exp & \checkmark & $\mathbb{E}_{p\sim Dir(\mathbf{\alpha})}\{\mathcal{H}[P(y|p)]\}, I[y,p] $ & \ding{55} \\ \midrule
				TPAMI2024~\cite{pei2024evidential} & Transfer Learning & Multi-Source-Free Unsupervised Domain Adaptation & $L_{ce-edl}$ & ReLU & \checkmark & $Sp=max_i e_i+1$ & \checkmark \\ \midrule
				AAAI2024~\cite{wang2024towards} & Supervised Learning & Open Set Object Detection & $L_{ce-edl} + L_{mse-edl}$ & Exp & \checkmark & $u=\frac{C}{S}$  & \checkmark \\ \midrule
				AAAI2024~\cite{xu2024reliable} & Supervised Learning & 3D  Mitochondria Segmentation & $L_{ce-edl}$ & Softplus & \checkmark & $u=\frac{C}{S}$  & \checkmark \\ \midrule
				ICML2024~\cite{pei2024evidential} & Transfer Learning & Few-Shot Open-Set Recognition & $L_{nll-edl}$ & Euclidean distance & \checkmark & $u=\frac{C}{S}$  & \checkmark \\ \midrule
				ICLR2024~\cite{yu2023uncertainty} & Graph Neural Networks & Hyperspectral Image Classification & $L_{nll-edl}$ & Exp/ReLU & \ding{55} & $u=\frac{C}{S}$  & \checkmark \\ \midrule
				CVPR2024~\cite{chen2024think} & Transfer Learning/Active Learning & Medical Image Analysis & $L_{ce-edl}$ & ReLU & \checkmark & $U_{\text{ale}}~ \&~ U_{\text{epi}} $  & \checkmark \\ \midrule
				CVPR2024~\cite{fan2024evidential} & Active Learning & Open-World Embodied Perception & $L_{nll-edl}$ & Exp & \checkmark & $u=\frac{C}{S}$  & \checkmark \\ 
				\bottomrule
			\end{tabular}
		}
		\vspace{0mm}
	\end{table*}

\subsubsection{Automatic Driving}
\label{AD}
Automatic driving, a safety-critical field, urgently requires reliable, interpretable, and efficient uncertainty estimation approaches. LiDAR, one of the most popular sensors in automatic driving, suffers from sparse data and environment-dependent sensor effects. To tackle these issues, Bauer et al.~\cite{bauer2019deep} employ evidential convolutional neural networks to incorporate sensor noise which can be extended to model uncertainty. Facing LiDAR-based odometry (LO), Ali et al.~\cite{ali2023delo} propose to jointly learn accurate frame-to-frame correspondences and model's predictive uncertainty as evidence to safe-guard LO predictions. Based on the two unique advantages of EDL, computational efficiency and evidential decision fusion, Liu et al.~\cite{liu2021efficient}  design a hybrid evidential fusion to achieve robust autonomous control.

Beyond LiDAR-based tasks, evidential theory is gaining attention in other automatic driving-related tasks, including monocular localization with 
high-definition (HD) maps~\cite{petek2022robust}, traffic forecasting~\cite{li2023distil}, trajectory prediction~\cite{cai2023evora,itkina2023interpretable}, pedestrian intention prediction~\cite{zhang2023trep}, and road segmentation~\cite{chang2022fast}. 
For the localization problem in urban scenarios with HD maps, Petek et al.~\cite{petek2022robust} develop evidential heads for semantic segmentation and object detection. 
For trajectory prediction, 
Itkina et al.~\cite{itkina2023interpretable} further divide the epistemic uncertainty into semantic concepts: past agent behavior, road structure map, and social context.
Additionally, Zhang et al.~\cite{zhang2023trep} 
employ EDL to reject model predictions of pedestrian intention. 
Road segmentation task provide two modalities, RGB and depth images, Chang et al.~\cite{chang2022fast} collect evidence in multiple scales for each modality and perform multimodal evidential fusion. 

\subsubsection{EDL in the Open World}
\label{OS}
The real-world scenarios are inherently open-ended and dynamic, compelling models to recognize and handle samples from previously unseen classes~\cite{scheirer2012toward}. Taking into account the outstanding performance of EDL in OOD detection, EDL demonstrates significant potential and unique advantages in open-world scenarios. 
Currently, there are two EDL directions for addressing open-world problems:
the first direction focuses on calibration, where network outputs are calibrated on known classes to learn compact representations, thereby distinguishing them from unknown samples~\cite{bao2021evidential,chen2023cascade,sapkota2023adaptive,bao2022opental,ji2024spectral,zhang2023learning,zhao2023open,sapkotameta, wang2024towards}; the second direction involves introducing unknown samples during training, directly penalizing them to increase the model's uncertainty regarding these samples~\cite{su2023hsic,zhu2022towards,yu2023adaptive}.

For calibration, three types of optimization-based strategies are widely-used. The first strategy is to add new regularizer to the original loss function~\cite{bao2021evidential,zhou2023trustworthy} or adopt meta training strategies~\cite{sapkotameta}. 
Bao et al.~\cite{bao2021evidential} propose to penalize the consistency the between predictions and uncertainty mass, enforcing a negative correlation between them.
The second strategy is to revise the original loss by sample re-weighting. Specifically, Bao et al.~\cite{bao2022opental} employ influence functions~\cite{koh2017understanding}, which measures the impact of the sample on the prediction outcome, to assign weights to the loss of each sample. Similarly, Sapkota and Yu~\cite{sapkota2023adaptive} combine Distributionally Robust Optimization (DRO)~\cite{namkoong2016stochastic} and Scheduler Functions (SF) to weight sample-specific mean square error loss. The third strategy~\cite{ji2024spectral} directly sharpens the decision boundary by additional samples generated by unsupervised models such as generative adversarial networks, variational autoencoders and normalizing flows.

While the aforementioned methods strictly adhere to open-set settings without making assumptions about unknown samples, the following methods incorporate unknown samples during training to enhance the distinction between known and unknown samples~\cite{su2023hsic,zhu2022towards,yu2023adaptive}. Su et al.~\cite{su2023hsic} treat highly uncertain samples as pseudo-unknown and encourage the dissimilarity between their probability and IoU scores. 
In addition, Yu et al.~\cite{yu2023adaptive} adopts the setting of open-set semi-supervised learning (Open-set SSL) where unknown samples are provided but unlabeled. 


\subsubsection{EDL for Science}
\label{SCI}
The application of evidential deep learning is gradually expanding to other scientific fields, including medicine, chemistry, physics, etc. 
In the medical field, the application of evidential networks precisely meets the high demands for reliability and trustworthiness, providing preliminary theoretical support for neural network-based automated medical diagnosis. While evidential application in the fields of physics and chemistry is relatively less common, it still demonstrates significant potential for future exploration.
{(1) Medicine.}
Currently, numerous studies consider using evidential theory as a robust tool for uncertainty quantification in radiograph classification to meet the stringent reliability requirements of medical tasks~\cite{gudhe2022multi, ghesu2019quantifying, gao2023reliable, hemmer2022deal, balaram2022consistency}. Additionally, several studies advance medical image classification to medical image segmentation, including  general medical image segmentation~\cite{xu2023eviprompt, ren2023uncertainty} and specialized brain tumor segmentation~\cite{zou2022tbrats, li2023region}. Beyond single-modality tasks, Zheng and Yu~\cite{zheng2023evidential} focus on evidence-enhanced medical image captioning. 
Last but not least, deep evidential regression~\cite{amini2020deep} is gaining traction in continuous glucose monitoring due to its inherited advantages of interpretability and efficiency from vanilla EDL~\cite{zhu2022personalized, zhu2022enhancing,zhu2022iomt}.
{(2) Physics.} Koh et al.~\cite{koh2021evaluating} assess the application of 
EDL on the task of particle classification in a simulated 3D LArTPC point cloud dataset. Tan et al.~\cite{tan2023single} examine evidential paradigm for improving the robustness of neural networks interatomic potentials through active learning.
{(3) Chemistry.} Soleimany et al.~\cite{soleimany2021evidential} leverage advances in evidential deep learning to form evidential 2D message passing neural networks and evidential 3D atomistic neural networks, targeted at quantitative structure-activity relationship regression tasks and key molecular discovery applications. Vazquez-Salazar et al.~\cite{vazquez2024outlier} apply deep evidential regression~\cite{amini2020deep} to detect samples with large expected errors, i.e. outliers, to reactive molecular potential energy surfaces.


\subsubsection{Discussion}
EDL won its popularity of various downstream tasks and subjects mainly for its {interpretability, simplicity, efficiency, generalizability}. 
{Interpretability} stems from DST and SL. These two theoretical foundations reinterpret the prediction of samples as a process of evidence collection, using the evidence to construct a Dirichlet distribution as the new prediction outcome, which enables the quantification of various types of uncertainty. 
{Simplicity} is reflected in the fact that transforming a neural network into an evidential network requires minimal modifications, often only to output layers and optimization objectives. As depicted in Table~\ref{tab_app}, simply adhering to the vanilla EDL which employs ReLU evidential function and NLL loss is empirically effective across diverse tasks. Simplicity also leads to another advantage, {Efficiency}, which refers to the fact that evidential networks require only a single forward pass to obtain uncertainty estimates, with significantly lower computational cost compared to Bayesian and ensemble methods. As a result, these three characteristics—interpretability, simplicity, and efficiency—endow EDL with applicability across various domains and tasks, i.e. {generalizability}.

However, every coin has two sides. Firstly, although it is possible to obtain evidence by changing only the output layer, this simple structure does not achieve satisfactory performance across all tasks. Therefore, the downstream applications of evidential networks are {context-dependent}. As shown in Tabel~\ref{tab_app}, additional context-dependent losses are often required to meet task settings. Conversely, the KL regularization is empirically not appropriate for most tasks. Moreover, while the formulation $u={C}/{S}$ is the most commonly used, context-dependent uncertainties are also significant. 
Secondly, EDL, in certain fields such as chemistry and physics~\cite{koh2021evaluating,soleimany2021evidential}, encounters a {paradox} in low computational cost and high quality of uncertainty estimation. Fortunately, evidential theory are advancing, and this issue is expected to be resolved. 
Thirdly, EDL's application scope awaits further {expansion}. While most works focus on computer vision-related tasks, EDL holds significant potential in other modalities, such as text and audio, as well as in other disciplines, such as geography and biology, due to its aforementioned characteristics.

\vspace{-1mm}
\section{Open Problems and Future Directions}
\label{sec:future}
\ParagraphB{Theoretical Enhancement and Analysis of EDL.} Subjective logic theory forms the cornerstone of deep evidence learning. Under the bijection condition between subjective opinions (\eg, binomial, multinomial~\cite{sensoy2018evidential}, and hypernomial~\cite{li2024hyper}) and Dirichlet distributions, a series of EDL methods have been developed. Subsequently, focusing on EDL based on improved subjective logic theory will be a feasible path. For instance, exploring the integration of subjective logic with other probabilistic models and reasoning frameworks, such as fuzzy logic~\cite{hajek2013metamathematics} and the Imprecise Dirichlet Model~\cite{walley1996inferences}, offers potential. Developing new operators guided by subjective logic, such as the fusion of multiple subjective opinions~\cite{han2022trusted}, is also a viable path. Furthermore, the theoretical guarantees of EDL are a worthy research direction. Previous work has partially explored the generalization error bounds of EDL using PAC learning theory~\cite{deng2023uncertainty, mcallester1998some}. Additionally, it is worth investigating the uncertainty estimation bounds~\cite{cerutti2022evidential,ulmer2021prior}, robustness bounds against adversarial examples, convergence and stability bounds of model training, and the information loss and compression bounds. Last but not least, what are the fundamental differences between the optimization objectives of EDL and popular loss functions such as cross-entropy and MSE?~\cite{chen2023r} further elucidated the effectiveness of EDL by relaxing its non-essential components, but a more comprehensive comparative analysis of these optimization objectives, such as gradient analysis and bias-variance analysis, would still be beneficial. Such analyses could further explain the potential of EDL and provide insights into the optimization of neural networks.

\ParagraphB{More Meaningful Evidence Collection.} Although the term ``evidence'' reflects the essence of evidential deep learning, current approaches often overlook the physical significance of why the evidence scores collected by neural networks can be considered as evidence. Given that neural networks are black-box models with mechanisms that are difficult to interpret, there exists a cognitive gap between the obtained evidence scores and the explicit belief required by subjective logic. Currently, some works~\cite{gao2023vectorized,chen2023cascade} have moved beyond merely using model output logits as sources of evidence, adopting more sophisticated evidence collection methods or exploring a broader range of evidence sources, there remains significant potential for further exploration in this area.
For instance, integrating additional external prior knowledge (\eg, knowledge graphs and causal inference) or drawing on principles from cognitive psychology could potentially enhance the EDL paradigm.

\ParagraphB{Better Uncertainty Estimation.} Currently, most existing taxonomies categorize uncertainties as either aleatoric or epistemic~\cite{hullermeier2019aleatoric}. Specifically speaking, epistemic uncertainty (\ie, model uncertainty) has direct correlations with the learned models, which can be generally reduced by collecting new data or refining models. Aleatoric uncertainty (\ie, data uncertainty) is originated from the inherent noise of data hence irreducible in most cases. 
In EDL, the simplest way to calculate epistemic uncertainty is to associate it inversely with the total amount of evidence (\eg, $u=C/S$). However, this approach is somewhat arbitrary as it neither originates from the fundamental principles of subjective logic nor often achieves significant effectiveness in practical applications~\cite{deng2023uncertainty,chen2023r}. For aleatoric uncertainty, there has been no mature method to adumbrate such inherent data properties without external tools, and the only viable way to obtain an indirect estimation is to analyze the model outputs. We acknowledge that such aleatoric uncertainty derived from model outputs will inevitably be affected by epistemic (model) uncertainty, but how to obtain ``pure'' aleatoric uncertainty still remains an open problem. In the future, how to estimate various forms of uncertainty in EDL in accordance with the principles of subjective logic theory is a direction worth exploring. Besides, although EDL requires minimal additional computational overhead, it is often criticized for its performance, which is weaker than mainstream uncertainty estimation algorithms. Integrating EDL with existing robust uncertainty quantification algorithms, such as deep ensemble~\cite{lakshminarayanan2017simple}, could potentially advance the field.

\ParagraphB{Enhancing pre-trained foundation models with EDL.} In recent years, pre-trained large models have demonstrated remarkable performance across various tasks~\cite{zhao2023survey, zhou2023comprehensive}.
According to the scaling law~\cite{zhao2023survey}, the key factors influencing the performance of foundational large models are model size, data quantity, and computational power. Even so, we cannot dismiss the potential of improving optimization objectives to enhance large model performance. In fact, EDL can be seamlessly integrated into the optimization process of large models. For example, in autoregressive token prediction, incorporating EDL can enable the model to consider higher-order probability distributions and uncertainty predictions. So far, possibly due to constraints such as computational power of 
research teams related to EDL, research on EDL has mainly focused on small models, with applications to large models yet to be explored, while this gap inspires further research. Furthermore, besides training and fine-tuning large models directly, leveraging EDL strategies for distillation~\cite{malinin2019ensemble} and compression of large models is also a viable direction.
Finally, a valuable research direction with industrial and high-risk applications is exploring whether the EDL paradigm can endow large models with uncertainty estimation capabilities without increasing computational overhead or compromising performance.

\ParagraphB{Broader applications.} As discussed in previous sections, EDL has already been explored in various fields, including computer vision, natural language processing, cross-modal learning, and scientific research. Given the advantages of EDL in uncertainty awareness and efficiency, further exploration in large-scale practical applications is valuable. For instance, in embodied AI, there is currently little to no related research to our knowledge~\cite{fan2024evidential}. Additionally, EDL is expected to achieve successful applications in content generation, such as improving the generation quality of Diffusion models~\cite{croitoru2023diffusion} based on probabilistic modeling.

\vspace{-1mm}
\section{Conclusion}
\label{sec:conclusion}
In this survey, we review the recent progress in evidential deep learning (EDL), elaborating on both theoretical advancements and real-world applications. 
Specifically, we discuss theoretical explorations of EDL in four aspects: reformulating the evidence collection process, improving uncertainty estimation via OOD samples, delving into various training strategies, and evidential regression networks.
Following this, we introduce EDL's extensive applications across various machine learning paradigms and real-world downstream tasks.
Additionally, we review subjective logic theory, the theoretical foundation of EDL, clearly demonstrating how researchers apply this theory in deep learning to develop the EDL method and discussing its distinctions from other uncertainty reasoning frameworks.
The definition and classification of the concept of uncertainty, as well as a comparison of different uncertainty estimation methods, are also discussed to deepen the readers' understanding of EDL.
Generally speaking, this work provides a comprehensive overview of current research on EDL, designed to offer readers a broad introduction to the field without assuming prior knowledge.
Additionally, this survey covers the most recent literature on EDL, serving as a valuable reference for both researchers and engineers.

{
\small
\bibliographystyle{IEEEtran}
\bibliography{reference}
}

%
%
%
%
%
%
%

\end{document}